\documentclass[11pt]{article}
\usepackage{subcaption}

\usepackage[preprint]{acl}

\usepackage{times}
\usepackage{latexsym}

\usepackage[T1]{fontenc}

\usepackage[utf8]{inputenc}

\usepackage{microtype}

\usepackage{inconsolata}

\usepackage{graphicx}

\usepackage{times,latexsym}
\usepackage{enumitem}
\usepackage{amsmath,amssymb,amsthm}
\usepackage{graphicx}
\usepackage{booktabs}
\usepackage{enumitem}
\usepackage{color}
\usepackage{algorithm}
\usepackage{algorithmic}
\usepackage[most]{tcolorbox}
\usepackage{microtype}
\usepackage{xcolor}
\usepackage{multirow}
\usepackage[table]{xcolor}
\usepackage{inconsolata}
\usepackage{threeparttable}
\usepackage{textcomp}
\usepackage[normalem]{ulem}
\usepackage[acronym]{glossaries}
\makeglossaries
\newacronym{ICL}{ICL}{In-Context Learning}
\newacronym{LLMs}{LLMs}{Large Language Models}
\newacronym{CoT}{CoT}{Chain-of-Thought}
\newacronym{SCOS}{SCOS}{Semantic Consistency Optimisation Sampling}

\usepackage{adjustbox}

\usepackage{amsthm}
\usepackage{threeparttable}
\usepackage{enumitem}

\usepackage[most]{tcolorbox}
\usepackage{listings}

\lstdefinestyle{promptstyle}{
  basicstyle=\ttfamily\small,
  breaklines=true,
  breakatwhitespace=false,
  columns=fullflexible,
}

\newtcblisting{promptbox}[1][]{
  listing only,
  breakable,
  enhanced,
  colback=gray!6,
  colframe=gray!60,
  boxrule=0.4pt,
  arc=2pt,
  left=1mm,right=1mm,
  top=1mm,bottom=1mm,
  listing options={style=promptstyle},
  colbacktitle=gray!12,      
  coltitle=black!70,         
  fonttitle=\bfseries\footnotesize,
  boxed title style={sharp corners},
  title=#1,
}

\usepackage[most]{tcolorbox}

\newtcolorbox{pro2}[1][]{
  breakable,
  enhanced,
  colback=black!2,
  colframe=black!18,
  boxrule=0.35pt,
  arc=1pt,
  left=1.2mm,
  right=1.2mm,
  top=0.8mm,
  bottom=0.8mm,
  colbacktitle=black!5,
  coltitle=black,
  fonttitle=\bfseries\footnotesize,
  title=#1,
  fontupper=\small,
  before upper=\setlength{\parindent}{0pt}\setlength{\parskip}{0.35em}\raggedright,
}

\title{Chain-of-Thought as a Lens: Evaluating Structured Reasoning Alignment between Human Preferences and Large Language Models}

\author{
 \textbf{Boxuan Wang},
 \textbf{Zhuoyun Li},
 \textbf{Xinmiao Huang},
 \textbf{Xiaowei Huang},
 \textbf{Yi Dong\thanks{Corresponding Author: yi.dong@liverpool.ac.uk}}
 \\
 School of Computer Science and Informatics, University of Liverpool, United Kingdom
 \\
   \small{\texttt{\{boxuan.wang, zhuoyun.li, xinmiao.huang, xiaowei.huang, yi.dong\}@liverpool.ac.uk}}
}

\begin{document}
\maketitle
\begin{abstract}
This paper primarily demonstrates a method to quantitatively assess the alignment between \emph{multi-step, structured reasoning} in large language models and human preferences. We introduce the Alignment Score, a semantic-level metric that compares a model-produced chain of thought traces with a human-preferred reference by constructing semantic-entropy-based matrices over intermediate steps and measuring their divergence. Our analysis shows that Alignment Score tracks task accuracy across models and hop depths, and peaks at 2-hop reasoning. Empirical results further indicate that misalignment at greater reasoning depths is driven mainly by alignment errors such as \emph{thematic shift} and \emph{redundant reasoning}. Viewing chain sampling as drawing from a distribution over reasoning paths, we empirically demonstrate a strong and consistent correlation between Alignment Score and accuracy, readability, and coherence, supporting its use as a diagnostic signal. The code is available.\footnote{\url{https://github.com/boxuanwang28/CoT-Lens}}

\end{abstract}

\section{Introduction}

\begin{figure}[t]
    \centering
    \hspace*{-10pt}
    \includegraphics[width=0.49\textwidth]{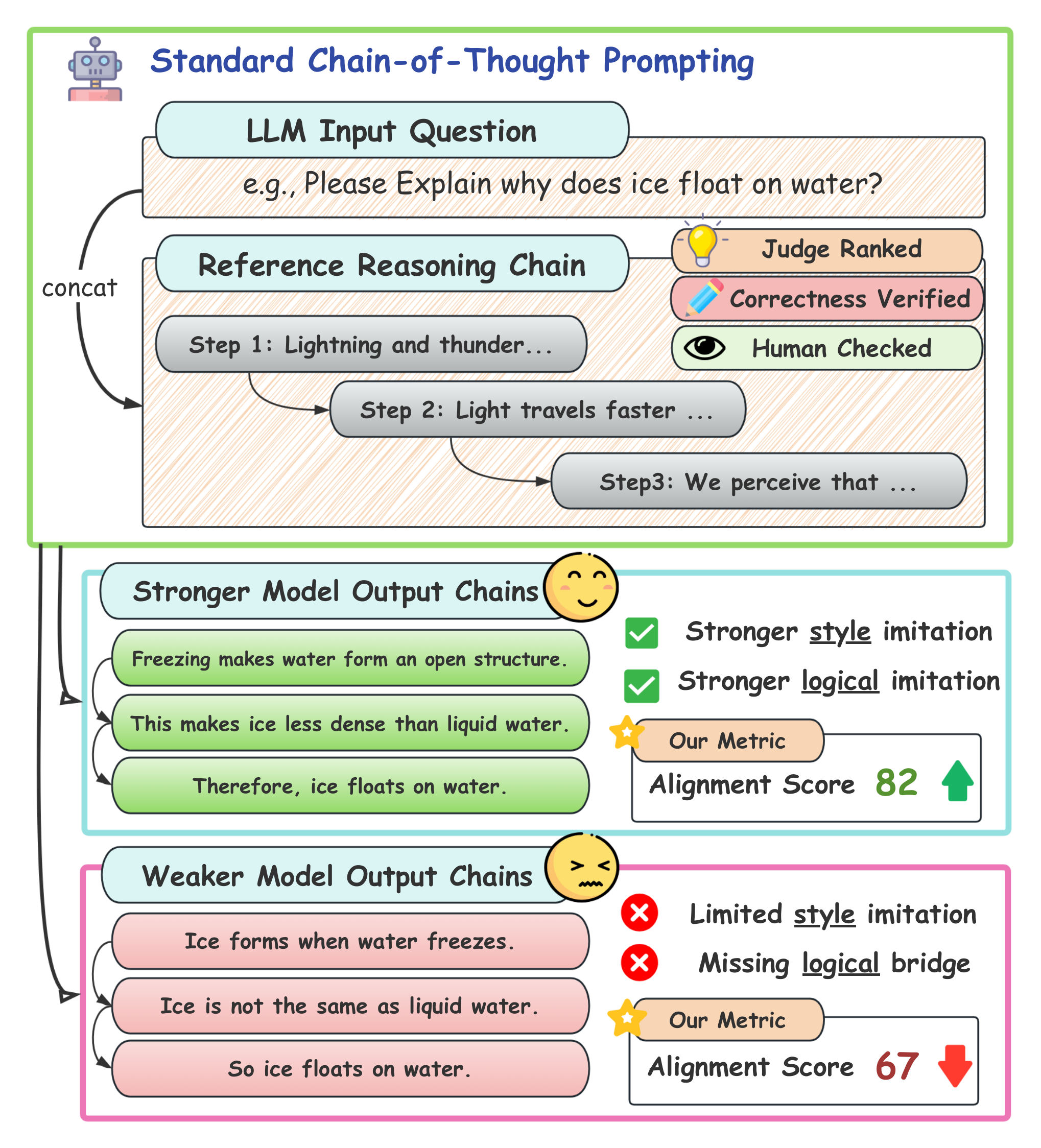}
    \caption{An illustration of comparing reasoning consistency in strong vs. weak models under CoT prompting: the proposed Alignment Score quantifies their divergence, with higher values indicating the stronger style imitation and improved logical coherence.}
    \label{fig:example}
\end{figure}

\begin{figure*}[t]
    \centering
    \includegraphics[width=1\textwidth]{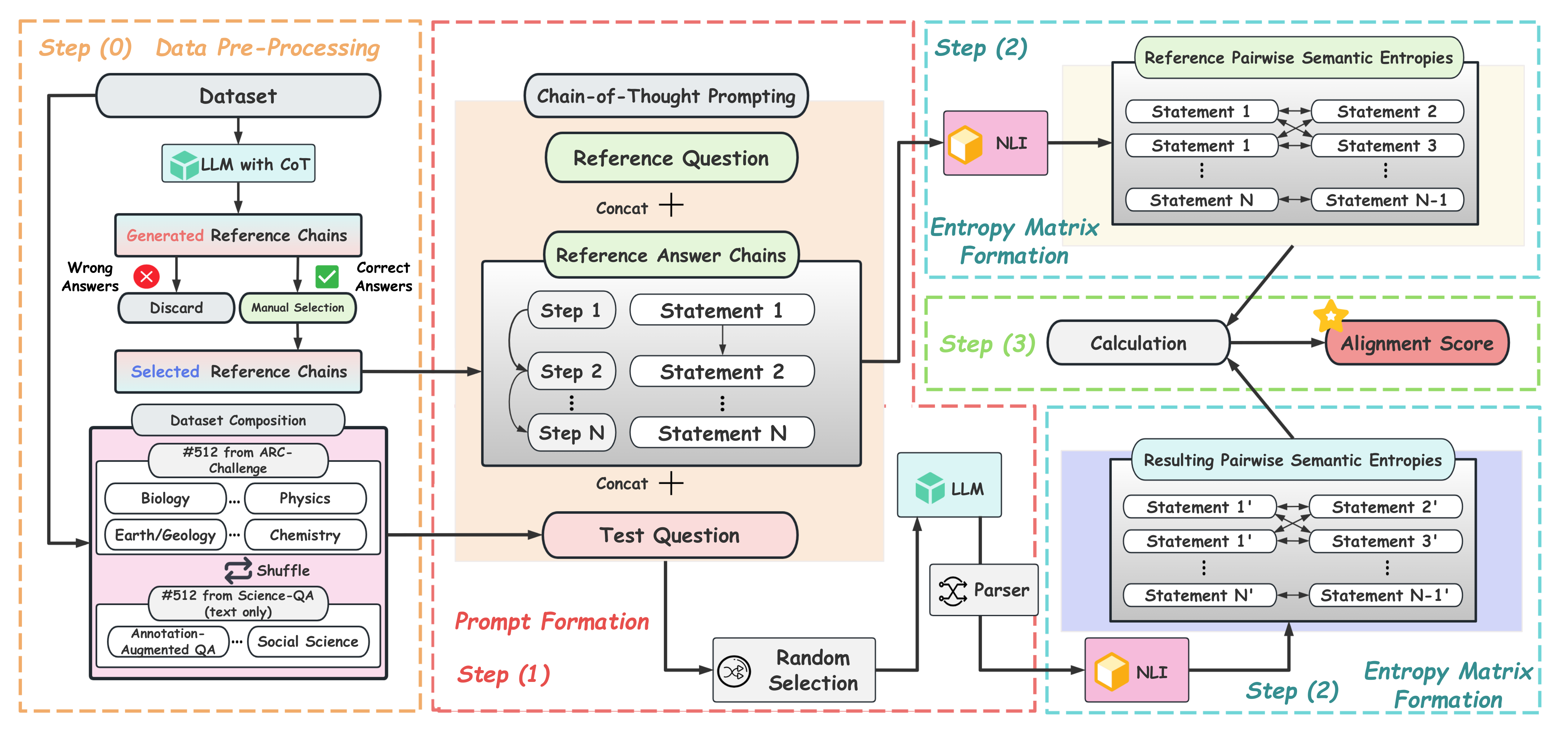}
    \caption{Illustration of the Alignment Score Computation: \textbf{\uline{Step (0)}}: Prepare the dataset and select reference chains (orange box). \textbf{\uline{Step (1)}}: Construct the prompt with a reference reasoning chain and generate model reasoning chains (red box; see Appendix \ref{app:prompts} for prompt templates); \textbf{\uline{Step (2)}}: Compute pairwise semantic entropy matrix using an NLI model (blue box); \textbf{\uline{Step (3)}}: Compare the resulting entropy matrix and derive the Alignment Score (green box).}
    \label{fig:framework}
\end{figure*}

In-Context Learning (ICL) is widely regarded as a core capability of Large Language Models (LLMs), allowing them to tackle diverse tasks by conditioning on a few example instances included in the input prompt, all without modifying their parameters during inference \cite{dong2022survey}. A particular approach within ICL is Chain-of-Thought (CoT) prompting \cite{wei2022cot}, which encourages models to generate intermediate reasoning steps to achieve a final answer. 
CoT has been shown to significantly enhance model performance in complex reasoning tasks, e.g., arithmetic reasoning~\cite{cobbe2021gsm8k}, symbolic logic~\cite{zhou2022least}, and commonsense question answering~\cite{kojima2023largelanguagemodelszeroshot}. Although CoT and its established variants (e.g., Least-to-most~\cite{zhou2022least} and self-consistency~\cite{wang2022selfconsistency}) have led to gains in both robustness and accuracy, recent work shows that LLM reasoning traces still vary substantially in quality~\cite{turpin2023language,hu2025stopreducingresponsibilityllmpowered}. In particular, structured or multi-step reasoning often suffers from semantic incoherence, logical inconsistency, or thematically misaligned steps, even when the final answer is correct \cite{lampinen2022can, kojima2023largelanguagemodelszeroshot}. These findings expose a key gap: we still lack an evaluation metric that goes beyond answer correctness and captures the quality of the reasoning process itself. 

As illustrated in Figure~\ref{fig:example}, to address this gap, we revisit multi-step reasoning in CoT through the perspective of \textit{reasoning consistency}.\footnote{\textit{Reasoning consistency} refers to the extent of logical congruence and step-by-step agreement between the model’s generated reasoning chains and human-preferred reference chains (i.e., coherent and readable traces with correct final answers).} Our central hypothesis is that poor reasoning performance often arises not from models being unable to reach correct conclusions, but from the fact that the reasoning chains they produce deviate from reference trajectories that reflect correct, coherent, and human-preferred logical structure.


Building on this motivation, we design an Alignment Score framework that treats CoT as the main handle for measuring how well model reasoning aligns with human-preferred chains, as shown in Figure~\ref{fig:framework}. We first manually filter model-generated CoT explanations to obtain a pool of correct, well-structured reference chains. These references are inserted as in-context exemplars for each test question, prompting the model to imitate their reasoning style. We then compare the reference chain and the generated chain in a latent space defined by semantic entropy~\cite{kuhn2023semantic}: pairwise judgments form entropy matrices over reasoning steps, and their divergence is summarized as a single Alignment Score that reflects structural and logical, rather than surface-level, agreement.

To interpret this score and link it to task performance, we analyze how specific alignment errors and reasoning depth affect both Alignment Score and accuracy. From empirical inspection, we define two interpretable error types, \textit{Thematic Shift} and \textit{Redundant Reasoning}, and track how their frequencies evolve with the number of reasoning hops. Finally, we introduce two alignment-aware sampling schemes, Alignment-aware Chain Sampling and Selection (ACSS) and Self-Consistency with Alignment (SC-Align), which sample multiple CoT chains under a fixed budget and then select chains using alignment-based criteria, allowing us to probe whether higher Alignment Scores correspond to higher accuracy and better reasoning quality. Our main contributions are threefold:
\begin{enumerate}[label=(\arabic*), leftmargin=*]
    \item We propose the \textbf{Alignment Score}, a metric that quantitatively measures how closely model-generated reasoning traces align with human-preferred reference chains, and validate its effectiveness and reliability through comparison-based correlation analysis.
    \item We conduct a detailed empirical study of how reasoning depth influences Alignment Score, and show that \textit{Thematic Shift} and \textit{Redundant Reasoning} emerge as dominant alignment errors as reasoning depth increases.
    \item We present \textbf{ACSS} and \textbf{SC-Align} as alignment-aware sampling methods, with two main findings: (i) Alignment Score closely tracks task accuracy; and (ii) Alignment Score is strongly associated with improved readability and coherence under extensive LLM-based evaluations.
\end{enumerate}

\section{Related Work}


\paragraph{LLM Evaluation Techniques.} Evaluating the quality of LLM outputs has long been a central topic. Existing answer‑focused benchmarks such as MMLU~\cite{hendrycks2021m}, ARC‑AGI~\cite{chollet2019measure}, GPQA~\cite{rein2024gpqa}, GSM8K~\cite{cobbe2021gsm8k}, TruthfulQA~\cite{lin2022truthfulqa}, and the broad BIG‑Bench suite~\cite{srivastava2022bigbench} together with its harder subset BBH~\cite{suzgun2022challenging}, cover a wide range of tasks but largely ignore whether \emph{reasoning process} is plausible or self‑consistent.  
To bridge this gap, several evaluation frameworks specifically targeting CoT have recently emerged. REVEAL~\cite{jacovi2024reveal} provides fine‑grained step‑level annotations to measure logical correctness. The CoTKG paradigm of Nguyen \emph{et~al.}~\cite{nguyen2024direct} directly grounds intermediate steps in knowledge graphs (KGs), producing discriminative and generative scores for multi‑hop reasoning. Vacareanu \emph{et~al.}~\cite{vacareanu2024general} further introduces a general-purpose verifier that checks each step for relevance, mathematical accuracy, and logical consistency. Our approach differs from the above by evaluating LLMs from the perspective of reasoning consistency between generated reasoning chains and reference CoT demonstrations. Unlike REVEAL and CoTKG, our evaluation method does not rely on step-level annotations, external verifiers, or knowledge graph construction, making it more lightweight and broadly applicable.

\paragraph{Semantic Entropy.} 
Semantic-level uncertainty methods aim to capture variability over \emph{meanings} rather than surface forms. 
Semantic Entropy has emerged as an effective approach for estimating LLM uncertainty by sampling outputs, clustering semantically equivalent responses via entailment, and computing entropy over the induced meaning distribution~\cite{farquhar2024detecting}. 
Building on this idea, Kernel Language Entropy (KLE) uses kernels over semantic similarities to obtain smoother entropy estimates~\cite{nikitin2024kernel}, Shapley Uncertainty decomposes semantic variability via Shapley-value attributions~\cite{zhu2025shapley}, and Semantic Energy combines Boltzmann-style energy with semantic clustering for confidence estimation~\cite{ma2025semantic}. 
In this work, we apply semantic entropy in a \emph{one-to-one} alignment setting, directly measuring semantic dispersion between each generated reasoning step and its corresponding reference step. 
To our knowledge, this constitutes the first systematic use of semantic entropy for evaluating \emph{step-structured reasoning} in LLMs.

\section{Methodology}

In this section, we first present the necessary preliminaries. Then, we present the calculation of the Alignment Score. Finally, we introduce the methods of alignment-aware chain selection.
\subsection{Preliminaries}

\paragraph{Pairwise Semantic Entropy}

Semantic entropy has been proposed as a measure of semantic uncertainty, practically based on Natural Language Inference (NLI) entailment signals,~\cite{kuhn2023semantic,farquhar2024detecting,vashurin-etal-2025-benchmarking,kossen2024semantic}. Given a pair of statements \( (S_i, S_j) \), the semantic relationship: entailment (E), contradiction (C), or neutrality (N), is predicted by an NLI model, which gives a probability distribution \( \{p_E, p_C, p_N\} \). The pairwise entropy \(H(S_i, S_j)\) can then be computed as:
\begin{equation}
H(S_i, S_j) = - \sum_{k \in \{E, C, N\}} p_k \log_2(p_k)
\end{equation}

While prior CoT evaluation frameworks~\cite{jacovi2024reveal,nguyen2024direct} typically assess local sentence pairs in isolation (e.g., an intermediate claim and its premise), our goal is different: we seek to evaluate the internal semantic coherence within a reasoning chain, as well as the semantic consistency between two chains. To this end, we propose a pairwise formulation of semantic entropy over all step pairs in a reasoning path.

In this scenario, a reasoning chain can be presented as \( \mathcal{C} = \{S_1, S_2, \dots, S_n\} \), where \( S_i \) is the \( i \)-th statement in the reasoning chain. 
Similarly, the reference reasoning chain can be denoted as \( \mathcal{C}^{\text{ref}} = \{S_1^{\text{ref}}, S_2^{\text{ref}}, \dots, S_n^{\text{ref}}\} \), where the superscript “ref” indicates the human-preferred reference chain. For each pair of statements \( (S_i, S_j) \) where \( i, j \in \{1, 2, \dots, n\} \) and \( i \ne j \), we compute the pairwise semantic entropy as:
\begin{equation}
H_{i,j} = H(S_i, S_j)
\end{equation}
For convenience, we represent the semantic entropy between two statements by \( H(i, j) \). This pairwise form enables us to quantify semantic consistency both within and between reasoning chains.

Although LLM outputs are not inherently structured as strict stepwise sequences, recent studies have shown that appropriate prompting strategies and ICL techniques~\cite{dong2022survey} can effectively elicit multi-step reasoning behaviors from LLMs. In our setting, we assume that the reasoning output has been formatted into a list of steps through CoT prompting with chain-annotated demonstrations. Given \( n \) reasoning steps, there are \( \binom{n}{2} = \frac{n(n-1)}{2} \) unique pairs to evaluate, forming the basis for constructing the semantic entropy.

\paragraph{Semantic Entropy Matrix}

The pairwise semantic entropy values can be organized in matrix form for further structural comparison. There will be two matrices: one for model-generated reasoning chains, and one for human-written reference chains, as illustrated in the two blue boxes of Figure~\ref{fig:framework}.

Given a model output reasoning chain \( \mathcal{C} = \{S_1, S_2, \dots, S_n\} \), we construct its semantic entropy matrix \( \mathbf{H} \) as:
\begin{equation}
\small
\mathbf{H} =
\begin{bmatrix}
0 & H_{1,2} & H_{1,3} & \cdots & H_{1,n} \\
H_{2,1} & 0 & H_{2,3} & \cdots & H_{2,n} \\
\vdots & \vdots & \ddots & \vdots & \vdots \\
H_{n,1} & H_{n,2} & \cdots & 0 \\
\end{bmatrix}
\end{equation}

Similarly, for the reference reasoning chain \( \mathcal{C}^{\text{ref}} = \{S_1^{\text{ref}}, S_2^{\text{ref}}, \dots, S_n^{\text{ref}}\} \), we can construct its corresponding semantic entropy matrix \( \mathbf{H}^{\text{ref}} \).


\subsection{Calculation of Alignment Score}
\label{subsec:alignment-calculation}

To measure how far the model’s reasoning structure deviates from the reference, we first compute a \emph{semantic divergence} $D_{\text{sem}}$ between their normalized entropy matrices. Concretely, we take the upper-triangular entries of the reference entropy matrix and the model entropy matrix $\mathbf{H}^{\text{ref}}$ and $\mathbf{H}$, normalize them into two probability distributions $\mathbf{p}^{\text{ref}}$ and $\mathbf{p}$. 
Here, $\mathrm{KL}(\cdot\|\cdot)$ denotes the Kullback--Leibler (KL) divergence between two probability distributions.
We then define:
\begin{equation}
  \small
  D_{\text{sem}}
  = \tfrac{1}{2}\,\mathrm{KL}\!\Bigl(\mathbf{p}^{\text{ref}} \,\Big\|\, \tfrac{\mathbf{p}^{\text{ref}} + \mathbf{p}}{2}\Bigr)
   + \tfrac{1}{2}\,\mathrm{KL}\!\Bigl(\mathbf{p} \,\Big\|\, \tfrac{\mathbf{p}^{\text{ref}} + \mathbf{p}}{2}\Bigr)
\end{equation}
We adopt the Jensen--Shannon (JS) divergence as our semantic divergence because it is bounded, and more numerically stable than plain KL divergence when the model assigns low probability mass to regions where the reference is confident~\cite{lin1991divergence,li2016diversity}.
A smaller $D_{\text{sem}}$ indicates that the model-induced semantic distribution $P$ is closer to the reference distribution $P^{\text{ref}}$ (i.e., smaller information mismatch) and thus reflects stronger step-wise semantic agreement and less misalignment. We then monotonically map $D_{\text{sem}}$ to a score in $[0,100]$, as detailed in Algorithm~\ref{alg:semantic_entropy}.

\begin{algorithm}[t]
\small
\caption{Alignment Score Calculation}
\label{alg:semantic_entropy}
\begin{algorithmic}[1]
\REQUIRE Dataset of questions with reference chains $\{\mathcal{R}_q\}$, model-generated chains $\{\mathcal{C}_q\}$, entropy encoder
\ENSURE Average Alignment Score $\bar{S} \in [0, 100]$
\STATE Initialize score list $\mathcal{S} \gets [\,]$
\FOR{each question $q$}
  \STATE Build reference entropy matrix $H^{\mathrm{ref}}$ from $\mathcal{R}_q$ and obtain normalized upper-triangular vector $\mathbf{p}^{\mathrm{ref}}$
  \FOR{each model chain $c \in \mathcal{C}_q$}
    \STATE Build entropy matrix $H$ from $c$ and obtain normalized upper-triangular vector $\mathbf{p}$
    \STATE Let $\mathbf{m} \gets (\mathbf{p}^{\mathrm{ref}} + \mathbf{p})/2$ and compute 
           $D_{\mathrm{sem}} \gets \tfrac{1}{2}\mathrm{KL}(\mathbf{p}^{\mathrm{ref}}\|\mathbf{m}) + \tfrac{1}{2}\mathrm{KL}(\mathbf{p}\|\mathbf{m})$
    \STATE $S \gets (1 - D_{\mathrm{sem}})\times 100$; append $S$ to $\mathcal{S}$
  \ENDFOR
\ENDFOR
\STATE \textbf{return} $\displaystyle \bar{S} = \frac{1}{|\mathcal{S}|} \sum_{S \in \mathcal{S}} S$
\end{algorithmic}
\end{algorithm}

\subsection{Alignment Errors}
\label{subsec:alignment-errors}

To better understand how our Alignment Score relates to concrete failure modes and downstream task performance, we explicitly define a small set of \emph{alignment errors} that are both quantifiable and easy to interpret. From empirical observations, two types of errors appear most frequently: \emph{Thematic Shift} (TS) and \emph{Redundant Reasoning} (RR). The two error types are defined as follows.

\paragraph{Thematic shift (TS).}
TS captures changes in topic or theme across adjacent reasoning steps. Significant thematic drift suggests that the model has lost coherence or deviated from the core reasoning path~\cite{golovneva2023roscoe}. We detect shifts by computing the cosine similarity between sentence embeddings of consecutive steps and counting how many times it falls below a threshold $\tau_{\mathrm{TS}}$. Let $e_{B_i}$ denote the embedding of the $i$-th step:
\begin{equation}
  \text{TS} = \sum_{i=1}^{n-1} 
  \mathbb{I} \bigl( \cos(e_{B_i}, e_{B_{i+1}}) < \tau_{\mathrm{TS}} \bigr).
\end{equation}

\paragraph{Redundant reasoning (RR).}
RR measures \emph{semantic} repetition in reasoning chains, rather than exact string duplication. High redundancy indicates that the model keeps “circling around” the same point instead of making genuine progress~\cite{chiang-lee-2024-reasoning,jang-etal-2025-verbosity}. Let $e_{B_i}$ be the embedding of step $B_i$. For each step, we check whether it is overly similar to \emph{any} previous step:
\begin{equation}
  \text{RR} = \sum_{i=2}^{n} 
  \mathbb{I} \Bigl(
    \max_{1 \le j < i} \cos(e_{B_i}, e_{B_j}) > \tau_{\text{RR}}
  \Bigr),
\end{equation}
where $\tau_{\text{RR}}$ is a high similarity threshold. A step is thus marked as redundant if its semantic content is almost entirely covered by an earlier step, even when the surface wording differs.

Thresholds $\tau_{\mathrm{TS}}$ and $\tau_{\mathrm{RR}}$ are selected via percentile-based calibration based on a development set. A summary can be found in Appendix~\ref{app:threshold}.

\subsection{Alignment-Aware Chain Selection}
\label{subsec:alignment-selection}

We use test-time compute to probe how task accuracy, Alignment Score, and alignment errors are related. For each question, we sample $K$ reasoning chains under the same decoding budget, and only vary how we select a single final chain. We propose the following two chain selection strategies:

\paragraph{Alignment-Aware Chain Sampling and Selection (ACSS).}
In the first strategy, we select one chain from the $K$ sampled chains following the \emph{Best-of-N}~\cite{10.5555/3495724.3495977} paradigm. Concretely, we either minimize the number of alignment errors (e.g., thematic shift or redundant reasoning) or directly maximize the Alignment Score over the $K$ chains. This simple scheme makes it easy to see how strongly these alignment signals correlate with downstream task accuracy.

\paragraph{Self-Consistency with Alignment (SC-Align).}
In the second strategy, we further ask whether chains that are more likely to give the correct answer naturally have higher Alignment Scores. To test this, SC-Align follows the \emph{Self-Consistency} \cite{wang2022selfconsistency} paradigm: we first apply majority voting over the $K$ chains to obtain the subset of chains that support this answer, then select the chain with the highest Alignment Score as the final explanation. This keeps the same test-time compute as self-consistency, while directly testing whether Alignment Score can serve as a practical handle for improving performance.

\subsection{LLM-based Preference Evaluation}
\label{subsec:llm_judge}

To check whether higher Alignment Scores match human preferences, we adopt an \emph{LLM-as-a-judge} framework~\cite{huang-etal-2025-empirical,bai-etal-2024-mt,liu-etal-2023-g} to minimize human subjectivity. For each question, the judge model is given the problem, answer options, the human reference chain, and two candidates: the baseline chains and the SC-Align chains. It assigns 1–10 scores for \emph{coherence} (logical completeness and consistency) and \emph{readability}, and outputs a pairwise preference for each aspect (which chain is better or tie) with brief explanations. Detailed configurations of the evaluation pipeline can be found in Appendix ~\ref{app:llm-eval-template}.

\newcommand{\textsmaller}[1]{{\scriptsize #1}}
\newcommand{\ms}[2]{#1~\textsmaller{(±#2)}}
\newcommand{\mname}[2]{#1~\textsmaller{(#2)}}
\newcommand{\hopone}[1]{\multicolumn{1}{|c}{#1}}   
\newcommand{\avgcol}[1]{\multicolumn{1}{c|}{#1}}  

\begin{table*}[t]
\centering

\small
\setlength{\tabcolsep}{2.9pt}
\renewcommand{\arraystretch}{1.2}

\begin{tabular}{@{\hspace{3pt}}l ccccc
  c@{\hspace{4pt}}c@{\hspace{3pt}}c
  >{\columncolor{white}}p{7pt}@{}}

\cmidrule[\heavyrulewidth]{1-9}
\multirow{2}{*}{\centering\textbf{Model Name}}
  & \multicolumn{5}{c}{\textbf{Hop-wise Alignment Score \& Avg.\ (Mean ± Std)}} 
  & \multicolumn{3}{c}{\textbf{Comparisons}} & \cellcolor{white} \\
\cmidrule(l{0pt}r{2pt}){2-6} \cmidrule(l{0pt}r{3pt}){7-9}
  & \hopone{\textbf{1-Hop}} & \textbf{2-Hop} & \textbf{3-Hop} & \textbf{4-Hop} & \avgcol{\textbf{Avg.}}
  & \textbf{MATH} & \textbf{GPQA} & \textbf{MMLU} & \cellcolor{white} \\
\cmidrule{1-9}

\multicolumn{9}{c}{\textit{Open-Source Models} {\small (non-reasoning)}} & \cellcolor{white} \\
\cmidrule{1-9}

\rowcolor[RGB]{245,245,245}
Falcon-7B-Instruct 
& \hopone{\ms{75.23}{26.94}} & \textbf{\ms{79.44}{16.01}} & \ms{73.38}{13.74} & \ms{70.44}{10.09} & \avgcol{\ms{76.02}{16.70}}
& 0.01 & 0.25 & 0.28 & \cellcolor{white} \\

\rowcolor[RGB]{245,245,245}
Qwen2.5-3B-Instruct 
& \hopone{\ms{75.93}{28.90}} & \textbf{\ms{81.64}{16.06}} & \ms{79.51}{12.86} & \ms{77.60}{11.77} & \avgcol{\ms{79.03}{17.40}}
& 0.76 & 0.30 & 0.65 & \cellcolor{white} \\

\rowcolor[RGB]{245,245,245}
LLaMA2-13B-Chat 
& \hopone{\ms{77.46}{27.56}} & \textbf{\ms{79.67}{17.06}} & \ms{77.19}{13.49} & \ms{72.28}{10.60} & \avgcol{\ms{78.10}{17.18}}
& 0.28 & 0.23 & 0.51 & \cellcolor{white} \\
\cmidrule{1-9}

\multicolumn{9}{c}{\textit{Closed-Source Models} {\small (non-reasoning)}} & \cellcolor{white} \\
\cmidrule{1-9}

\rowcolor[RGB]{238,245,252}
\mname{GPT-3.5-Turbo}{2023-11}
& \hopone{\ms{76.97}{28.09}} & \textbf{\ms{81.00}{16.59}} & \ms{78.50}{12.91} & \ms{77.07}{10.39} & \avgcol{\ms{78.82}{17.00}}
& 0.43 & 0.31 & 0.70 & \cellcolor{white} \\

\rowcolor[RGB]{238,245,252}
\mname{GPT-4o}{2025-03}
& \hopone{\ms{78.78}{26.43}} & \textbf{\ms{81.59}{16.23}} & \ms{78.81}{12.94} & \ms{74.59}{10.95} & \avgcol{\ms{79.73}{16.64}}
& 0.77 & 0.54 & 0.86 & \cellcolor{white} \\

\rowcolor[RGB]{238,245,252}
\mname{GPT-4o-Mini}{2024-11}
& \hopone{\textbf{\ms{81.58}{24.99}}} & \ms{80.66}{17.18} & \ms{77.99}{13.04} & \ms{74.80}{10.93} & \avgcol{\ms{80.08}{16.54}}
& 0.70 & 0.40 & 0.82 & \cellcolor{white} \\
\cmidrule{1-9}

\multicolumn{9}{c}{\textit{Reasoning Models}} & \cellcolor{white} \\
\cmidrule{1-9}

\rowcolor[RGB]{250,244,234}
\mname{GPT-o1}{2024-09}
& \hopone{\ms{80.20}{25.91}} & \textbf{\ms{82.85}{15.81}} & \ms{79.95}{11.60} & \ms{78.60}{10.22} & \avgcol{\ms{81.00}{15.89}}
& 0.96 & 0.78 & 0.92 & \cellcolor{white} \\

\rowcolor[RGB]{250,244,234}
\mname{DeepSeek\textendash R1}{2025-05}
& \hopone{\ms{77.30}{27.82}} & \textbf{\ms{87.82}{10.17}} & \ms{86.91}{12.70} & \ms{82.87}{11.06} & \avgcol{\ms{84.01}{15.44}}
& 0.97 & 0.72 & 0.91 & \cellcolor{white} \\
\cmidrule{1-9}

\multicolumn{6}{c}{\textbf{Pearson Correlation (w.r.t.\ Avg.\ Score)}} 
& 0.88 & 0.83 & 0.87 & \cellcolor{white} \\
\cmidrule[\heavyrulewidth]{1-9}
\end{tabular}

\caption{Hop-wise Alignment Score and comparison with other benchmarks. \textbf{Bold} values indicate the highest Alignment Score for each model. The benchmark scores in the ``Comparisons'' columns are normalized to the $[0,1]$ range for the ease of comparison. The date in parentheses indicates the assessed version of closed-source models.}
\label{tab:comparison_with_other_metrics}
\end{table*}

\definecolor{cotblock}{RGB}{245,245,245}    
\definecolor{cssblock}{RGB}{238,245,252}    
\definecolor{alignblock}{RGB}{250,244,234}  

\begin{table*}[t]
\centering
\small
\setlength{\tabcolsep}{1.35pt}
\renewcommand{\arraystretch}{1.15}

\newcommand{\mth}[1]{\makebox[2.7em][l]{\scriptsize #1}} 

\begin{tabular}{@{}l c c c c c c c c@{}}
\toprule

& \multicolumn{3}{c}{\textbf{Open-Source Models} {\scriptsize (non-reasoning)}}
& \multicolumn{3}{c}{\textbf{Closed-Source Models} {\scriptsize (non-reasoning)}}
& \multicolumn{2}{c}{\textbf{Reasoning Models}} \\
\cmidrule(lr){2-4} \cmidrule(lr){5-7} \cmidrule(lr){8-9}

\textbf{Method}
& \scriptsize\textbf{Falcon-7B}
& \scriptsize\textbf{Qwen2.5-3B}
& \scriptsize\textbf{LLaMA2-13B}
& \scriptsize\textbf{GPT-3.5-turbo}
& \scriptsize\textbf{GPT-4o-mini}
& \scriptsize\textbf{GPT-4o}
& \scriptsize\textbf{GPT-o1}
& \scriptsize\textbf{DeepSeek-R1} \\
\midrule

\multicolumn{9}{c}{\textit{Accuracy (mean $\pm$ std, \%)}} \\
\midrule

\rowcolor{cotblock}
\mth{CoT}
& $23.02\;\text{\scriptsize($\pm$1.39)}$
& $75.59\;\text{\scriptsize($\pm$2.30)}$
& $59.75\;\text{\scriptsize($\pm$3.90)}$
& $74.58\;\text{\scriptsize($\pm$2.53)}$
& $86.93\;\text{\scriptsize($\pm$1.07)}$
& $92.68\;\text{\scriptsize($\pm$0.60)}$
& $\mathbf{\textcolor{red}{96.12}\;\text{\scriptsize($\pm$0.12)}}$
& $\mathbf{\textcolor{red}{98.88}\;\text{\scriptsize($\pm$0.07)}}$ \\

\rowcolor{cotblock}
\mth{SC-CoT}
& $23.07\;\text{\scriptsize($\pm$0.99)}$
& $\mathbf{\textcolor{red}{77.71}\;\text{\scriptsize($\pm$1.80)}}$
& $61.39\;\text{\scriptsize($\pm$2.00)}$
& $76.91\;\text{\scriptsize($\pm$1.41)}$
& $88.16\;\text{\scriptsize($\pm$2.89)}$
& $93.02\;\text{\scriptsize($\pm$0.43)}$
& $93.61\;\text{\scriptsize($\pm$1.10)}$
& $96.36\;\text{\scriptsize($\pm$0.12)}$ \\

\rowcolor{cssblock}
\mth{ACSS-TS}
& $23.17\;\text{\scriptsize($\pm$2.19)}$
& $75.59\;\text{\scriptsize($\pm$2.30)}$
& $60.24\;\text{\scriptsize($\pm$1.81)}$
& $\mathbf{\textcolor{blue}{77.08}\;\text{\scriptsize($\pm$2.88)}}$
& $88.21\;\text{\scriptsize($\pm$2.62)}$
& $92.78\;\text{\scriptsize($\pm$0.59)}$
& $93.56\;\text{\scriptsize($\pm$2.82)}$
& $96.30\;\text{\scriptsize($\pm$0.20)}$ \\

\rowcolor{cssblock}
\mth{ACSS-RR}
& $22.50\;\text{\scriptsize($\pm$2.60)}$
& $73.78\;\text{\scriptsize($\pm$2.64)}$
& $60.11\;\text{\scriptsize($\pm$1.60)}$
& $75.51\;\text{\scriptsize($\pm$1.98)}$
& $87.10\;\text{\scriptsize($\pm$2.42)}$
& $92.94\;\text{\scriptsize($\pm$0.78)}$
& $94.73\;\text{\scriptsize($\pm$1.28)}$
& $96.49\;\text{\scriptsize($\pm$0.18)}$ \\

\rowcolor{cssblock}
\mth{ACSS-Ali.}
& $\mathbf{\textcolor{red}{25.21}\;\text{\scriptsize($\pm$3.05)}}$
& $76.00\;\text{\scriptsize($\pm$0.11)}$
& $\mathbf{\textcolor{red}{61.78}\;\text{\scriptsize($\pm$4.18)}}$
& $75.81\;\text{\scriptsize($\pm$4.38)}$
& $\mathbf{\textcolor{red}{89.92}\;\text{\scriptsize($\pm$1.03)}}$
& $\mathbf{\textcolor{red}{94.59}\;\text{\scriptsize($\pm$1.71)}}$
& $\mathbf{\textcolor{blue}{94.92}\;\text{\scriptsize($\pm$1.40)}}$
& $\mathbf{\textcolor{blue}{96.53}\;\text{\scriptsize($\pm$0.40)}}$\\

\rowcolor{alignblock}
\mth{SC-Align}
& $\mathbf{\textcolor{blue}{24.02}\;\text{\scriptsize($\pm$2.83)}}$
& $\mathbf{\textcolor{blue}{77.24}\;\text{\scriptsize($\pm$1.99)}}$
& $\mathbf{\textcolor{blue}{61.58}\;\text{\scriptsize($\pm$1.75)}}$
& $\mathbf{\textcolor{red}{77.40}\;\text{\scriptsize($\pm$3.31)}}$
& $\mathbf{\textcolor{blue}{88.81}\;\text{\scriptsize($\pm$2.75)}}$
& $\mathbf{\textcolor{blue}{93.02}\;\text{\scriptsize($\pm$0.64)}}$
& $94.82\;\text{\scriptsize($\pm$1.67)}$
& $96.36\;\text{\scriptsize($\pm$0.21)}$ \\

\midrule
\multicolumn{9}{c}{\textit{Alignment Score (mean $\pm$ std)}} \\
\midrule

\rowcolor{cotblock}
\mth{CoT}
& $73.38\;\text{\scriptsize($\pm$13.75)}$
& $79.18\;\text{\scriptsize($\pm$12.46)}$
& $77.19\;\text{\scriptsize($\pm$13.49)}$
& $78.50\;\text{\scriptsize($\pm$12.91)}$
& $78.18\;\text{\scriptsize($\pm$13.58)}$
& $78.81\;\text{\scriptsize($\pm$12.94)}$
& $79.95\;\text{\scriptsize($\pm$11.60)}$
& $86.91\;\text{\scriptsize($\pm$12.70)}$ \\

\rowcolor{cotblock}
\mth{SC-CoT}
& $75.84\;\text{\scriptsize($\pm$13.44)}$
& $79.47\;\text{\scriptsize($\pm$12.36)}$
& $76.70\;\text{\scriptsize($\pm$12.98)}$
& $77.61\;\text{\scriptsize($\pm$13.64)}$
& $78.60\;\text{\scriptsize($\pm$13.16)}$
& $78.82\;\text{\scriptsize($\pm$13.13)}$
& $79.02\;\text{\scriptsize($\pm$11.76)}$
& $86.15\;\text{\scriptsize($\pm$10.46)}$ \\

\rowcolor{cssblock}
\mth{ACSS-TS}
& $74.89\;\text{\scriptsize($\pm$13.30)}$
& $79.18\;\text{\scriptsize($\pm$12.46)}$
& $76.73\;\text{\scriptsize($\pm$13.57)}$
& $77.72\;\text{\scriptsize($\pm$13.29)}$
& $77.96\;\text{\scriptsize($\pm$13.48)}$
& $78.94\;\text{\scriptsize($\pm$13.28)}$
& $76.45\;\text{\scriptsize($\pm$14.99)}$
& $87.62\;\text{\scriptsize($\pm$14.58)}$ \\

\rowcolor{cssblock}
\mth{ACSS-RR}
& $77.52\;\text{\scriptsize($\pm$14.09)}$
& $79.48\;\text{\scriptsize($\pm$12.39)}$
& $76.93\;\text{\scriptsize($\pm$13.01)}$
& $78.22\;\text{\scriptsize($\pm$13.04)}$
& $78.34\;\text{\scriptsize($\pm$13.08)}$
& $79.09\;\text{\scriptsize($\pm$13.09)}$
& $80.21\;\text{\scriptsize($\pm$10.12)}$
& $87.74\;\text{\scriptsize($\pm$14.35)}$ \\

\rowcolor{cssblock}
\mth{ACSS-Ali.}
& $\mathbf{\textcolor{blue}{80.15}\;\text{\scriptsize($\pm$13.08)}}$
& $\mathbf{\textcolor{blue}{88.28}\;\text{\scriptsize($\pm$7.85)}}$
& $\mathbf{\textcolor{red}{86.71}\;\text{\scriptsize($\pm$8.82)}}$
& $\mathbf{\textcolor{red}{88.05}\;\text{\scriptsize($\pm$7.91)}}$
& $\mathbf{\textcolor{red}{87.45}\;\text{\scriptsize($\pm$8.50)}}$
& $\mathbf{\textcolor{blue}{87.90}\;\text{\scriptsize($\pm$8.37)}}$
& $\mathbf{\textcolor{red}{88.15}\;\text{\scriptsize($\pm$7.46)}}$
& $\mathbf{\textcolor{blue}{89.42}\;\text{\scriptsize($\pm$12.37)}}$ \\

\rowcolor{alignblock}
\mth{SC-Align}
& $\mathbf{\textcolor{red}{80.97}\;\text{\scriptsize($\pm$12.61)}}$
& $\mathbf{\textcolor{red}{88.33}\;\text{\scriptsize($\pm$7.65)}}$
& $\mathbf{\textcolor{blue}{86.00}\;\text{\scriptsize($\pm$8.92)}}$
& $\mathbf{\textcolor{blue}{87.34}\;\text{\scriptsize($\pm$8.78)}}$
& $\mathbf{\textcolor{blue}{86.79}\;\text{\scriptsize($\pm$9.22)}}$
& $\mathbf{\textcolor{red}{88.07}\;\text{\scriptsize($\pm$8.36)}}$
& $\mathbf{\textcolor{blue}{86.40}\;\text{\scriptsize($\pm$9.90)}}$
& $\mathbf{\textcolor{red}{89.58}\;\text{\scriptsize($\pm$10.33)}}$ \\

\bottomrule
\end{tabular}
\caption{Cross-model comparison of methods under 3-hop reasoning. Rows with gray, blue, and beige backgrounds correspond to the baselines, ACSS methods and SC-Align, respectively. ACSS-TS, ACSS-RR, and ACSS-Ali. denote chain selection methods that, respectively, minimize thematic shift/redundant reasoning, or maximize the Alignment Score. \textbf{\textcolor{red}{Red}} numbers indicate the best result for each model, and \textbf{\textcolor{blue}{blue}} numbers the second best. }
\label{tab:cross_model_grouped}
\end{table*}

\section{Experiments}



This section presents a series of experiments to answer the following research questions (RQs):

\begin{enumerate}[label=\textbf{RQ\arabic*:}, leftmargin=*]
    \item Is the Alignment Score a reliable metric for evaluating reasoning alignment?
    \item How does the depth of reasoning (i.e., reasoning hop) affect the Alignment Score?
    \item How strongly is the Alignment Score correlated with task performance (accuracy)?
\end{enumerate}

In the following sections, we first describe the experimental settings and how the reference chains are constructed, and then address each RQ in turn.

\subsection{Experimental Setup}

\paragraph{Dataset.}
We carry out our experiments on a fused evaluation set of 1{,}024 multiple-choice questions drawn from two benchmarks: the text-only subset of ScienceQA~\cite{lu2022learn} and the ARC-Challenge benchmark~\cite{Clark2018ThinkYH}. 


\paragraph{Models.}
Our experiments consider LLMs spanning open-source, closed-source, and reasoning-oriented families, covering parameter scales from a few billion to the hundred-billion level.
The open-source models are Falcon-7B-Instruct~\cite{tii2023falconllm}, Qwen2.5-3B-Instruct~\cite{qwen2_5_2025}, and LLaMA2-13B-Chat~\cite{touvron2023llama}. 
The closed-source non-reasoning models are GPT-3.5-turbo~\cite{openai2023gpt35turbo}, GPT-4o-mini~\cite{openai2024gpt4omini}, and GPT-4o~\cite{openai2024gpt4o}, accessed via the OpenAI API. 
The reasoning models are GPT-o1~\cite{openai2024o1} and DeepSeek-R1~\cite{deepseek2025r1}. Hyperparameter settings can be found in Appendix~\ref{subsec:hparams}.

\paragraph{Statistical Reporting.}
All experiments report the mean and standard deviation over five single runs.

\paragraph{Reference Chains Construction.} To evaluate semantic alignment, we construct 1{,}024 reference reasoning chains of varying depths (from 1-hop to 4-hop, 256 for each hop), where intermediate steps lead to the conclusion. We limit our evaluation from 1-hop to 4-hop reasoning for both practical and cognitive reasons. Empirical studies such as HotpotQA~\cite{yang2018hotpotqa} show that most multi-hop questions need only 1–2 supporting facts, and reasoning beyond 4 hops is rare in real few-shot settings. This practical limitation is further reinforced by cognitive constraints: human working memory typically holds only 3–4 "chunks"~\cite{cowan2001magical}, making longer reasoning chains more difficult to interpret and construct as reliable reference chains. Detailed process for reference chain construction can be found in Appendix ~\ref{app:ref-chains}.

\subsection{RQ1: Effectiveness of Alignment Score}

To assess the effectiveness of the proposed Alignment Score, we conduct a comparative evaluation across open-source, closed-source, and reasoning-oriented models. Table~\ref{tab:comparison_with_other_metrics} reports hop-wise Alignment Scores (from 1-hop to 4-hop) and their average, together with three normalized comparing benchmark scores, including MATH~\cite{hendrycks2021measuringmath}, GPQA~\cite{rein2024gpqa}, and MMLU~\cite{hendrycks2021m}.

\begin{figure}[t]
  \centering
  \includegraphics[width=\linewidth]{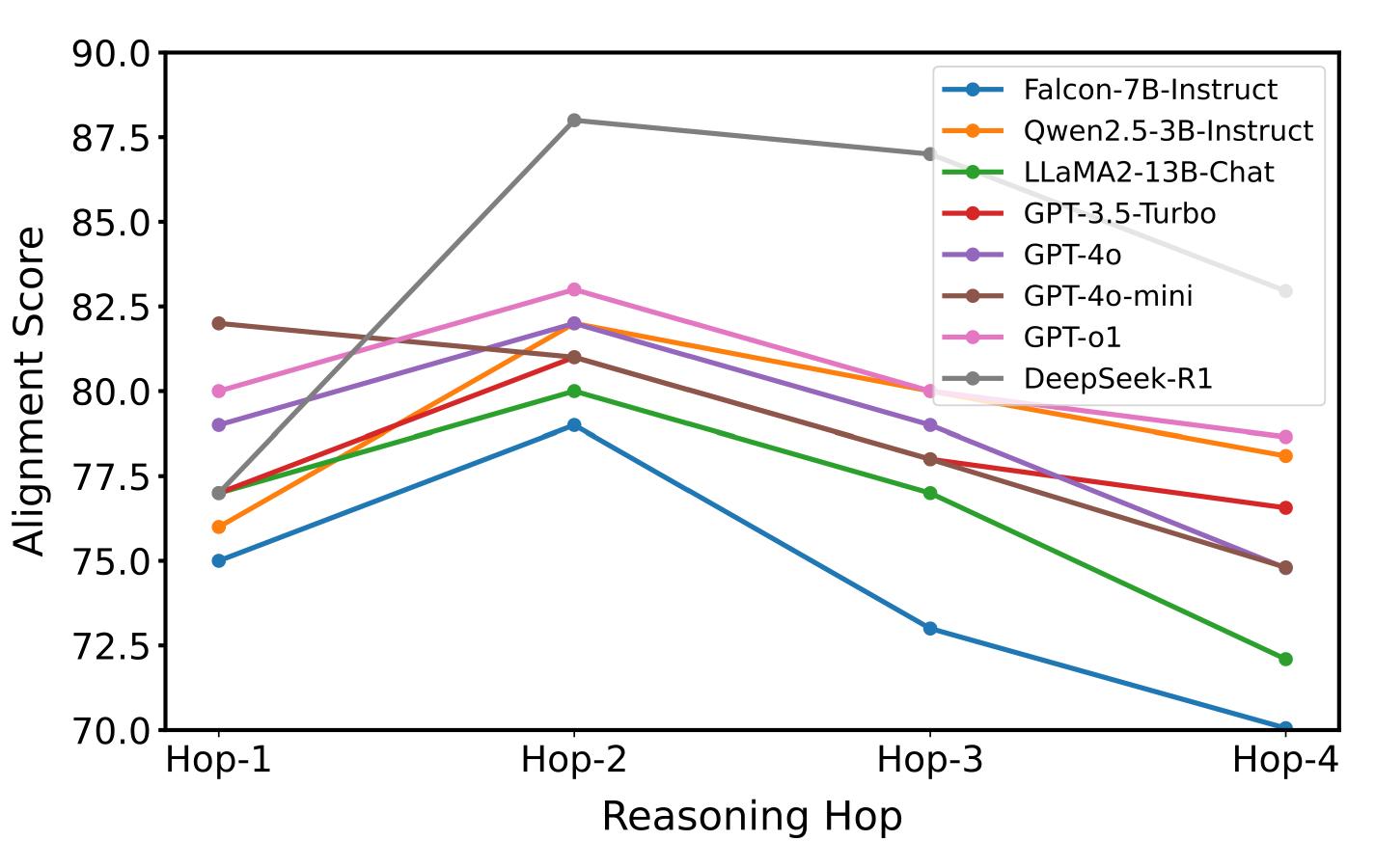}
  \caption{The Alignment Score trend across reasoning hops. Most models peak at 2-hop reasoning.}
  \label{fig:alignment-hop}
\end{figure}

\paragraph{Overall Trend.} A clear cross-model trend emerges, as illustrated in Figure~\ref{fig:alignment-hop}. Within the open-source group, Qwen2.5-3B-Instruct attains the highest average Alignment Score, followed closely by LLaMA2-13B-Chat and then Falcon-7B-Instruct. In the closed-source non-reasoning group, GPT-4o-mini and GPT-4o both substantially outperform GPT-3.5-turbo. Reasoning-oriented models (GPT-o1 and DeepSeek-R1) achieve the highest Alignment Scores overall. This ordering closely mirrors the commonly observed capability hierarchy of these models, indicating that the Alignment Score is sensitive to genuine reasoning strength.

\paragraph{Correlation with Comparison Benchmarks.} To further validate this, we compute Pearson correlations between the \emph{average} Alignment Score and comparison benchmarks. The Alignment Score correlates strongly with MATH (\textbf{0.88}), GPQA (\textbf{0.83}), and MMLU (\textbf{0.87}), which respectively target mathematical problem solving, graduate-level scientific reasoning, and broad multi-domain knowledge. Although Alignment Score is computed purely from reference-chain alignment rather than answer accuracy, it tracks benchmark performance closely and thus indicates that it can provide a reliable, task-agnostic signal of overall reasoning quality.

\begin{figure}[t]
  \centering
  \includegraphics[width=\linewidth]{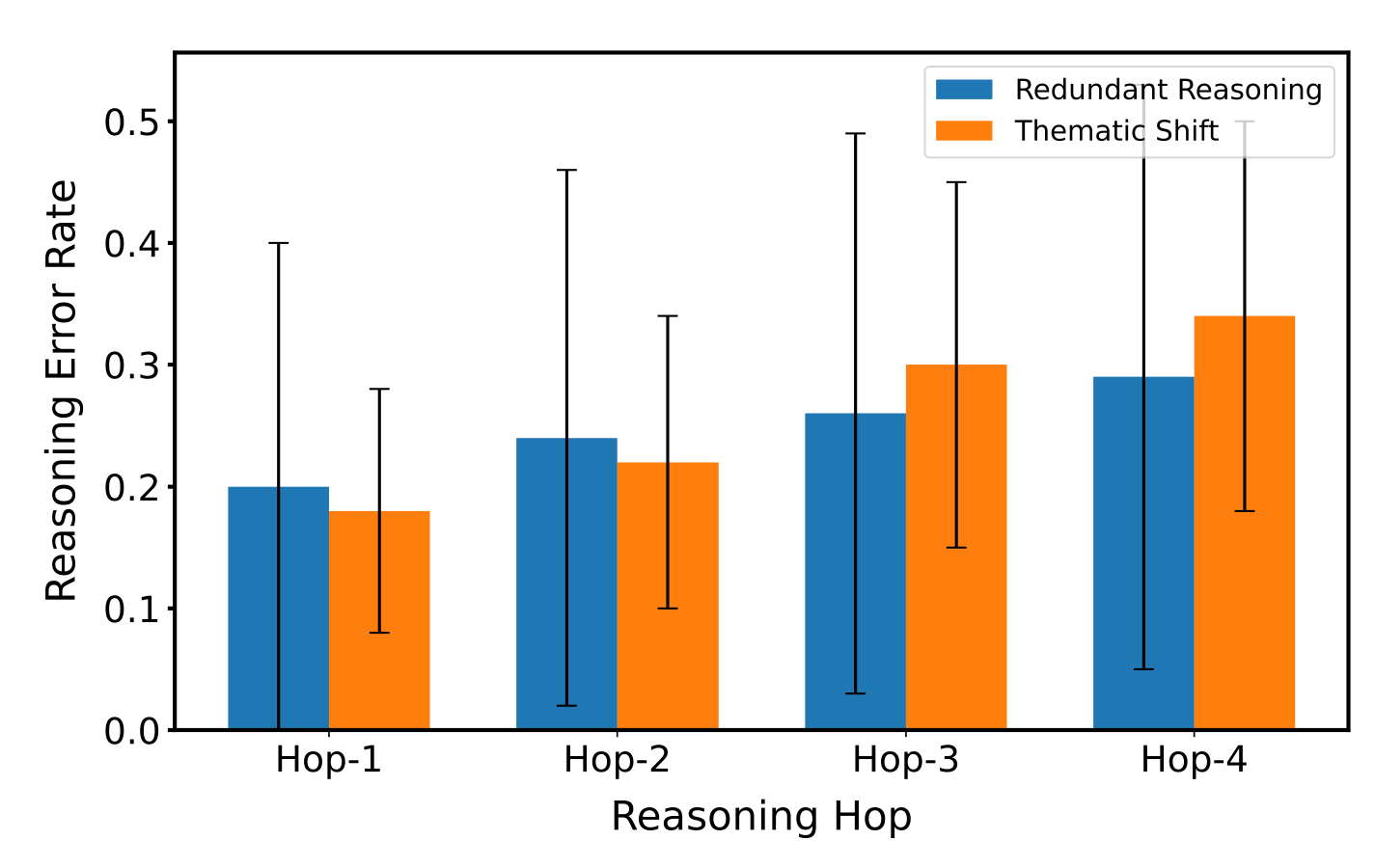}
  \caption{Alignment errors across reasoning hops.}
  \label{fig:error-hop}
  \vspace{-1em}
\end{figure}

\subsection{RQ2: Alignment Score Peaks at 2-Hop}
\label{sec:rq2_peak}

As illustrated in Figure~\ref{fig:alignment-hop}, we observe that the Alignment Score peaks at 2-hop reasoning in almost all models. In the following, we provide a theoretical and empirical explanation based on the structure of the semantic entropy matrix used in our metric.


\paragraph{1-hop is structurally under-informative.}
Recall that a reasoning chain $\mathcal{C} = \{S_1,\dots,S_n\}$ induces a semantic entropy matrix $\mathbf{H}$ (and $\mathbf{H}^{\text{ref}}$ for the reference chain), whose upper-triangular entries are normalized into probability distributions $\mathbf{p}$ and $\mathbf{p}^{\text{ref}}$; the semantic divergence $D_{\text{sem}}$ and the Alignment Score are then computed from these distributions (Section~\ref{subsec:alignment-calculation}). When reasoning hop $h=1$, both $\mathbf{p}$ and $\mathbf{p}^{\text{ref}}$ collapse to one-dimensional distribution supported by a single entropy value. In this degenerate scenario, $D_{\text{sem}}$ reduces to a single pairwise relation and contains no information about inter-step structure. As a result, the Alignment Score for 1-hop chains is intrinsically less sensitive to reasoning quality, because there is insufficient structural evidence to expose semantic misalignment. This justifies treating Alignment Score at 1-hop as a trivial lower-bound regime rather than a reliable indicator of multi-step reasoning ability, and explains why the Alignment Scores at 1-hop are relatively lower in most cases.

\paragraph{Deeper hops reduce alignment.}
For $h\ge2$, the semantic entropy matrix becomes expressive, enabling more meaningful comparison.
However, as hop length increases, the number of pairwise relations grows quadratically ($O(h^2)$), and reasoning errors propagate to multiple matrix entries.
Consequently, accumulated errors such as thematic drift and redundancy are amplified, leading to a systematic decrease in the score.
This effect is empirically reflected in the increased proportion of alignment errors at greater hops, as illustrated in Figure~\ref{fig:error-hop}.

\paragraph{Trade-off and 2-hop optimality.}
Taken together, 2-hop reasoning represents an effective trade-off: it is the minimal depth that yields a non-degenerate semantic structure while remaining short enough to limit error accumulation.
This balance explains the observed peak of the Alignment Score at 2-hop.

\subsection{RQ3: Does Alignment Score Predict and Improve Task Performance?}

Table~\ref{tab:cross_model_grouped} summarizes the results of the cross-model in terms of the accuracy of the task and the Alignment Score, comparing the CoT/SC-CoT baselines with the ACSS family of selectors and SC-Align. We explore whether the proposed Alignment Score is meaningfully related to task performance and whether it can be exploited as a test-time signal for chain selection. We study this question empirically using the ACSS family of selectors (which rank chains by alignment errors or the Alignment Score) and the SC-Align strategy (which orders chains by Alignment Score based on majority voting). Further experiments conducted on more recent-released models are presented in Appendix~\ref{app:additional_models}.

\begin{table}[t]
\centering
\small
\setlength{\tabcolsep}{6pt}
\renewcommand{\arraystretch}{0.95}
\begin{tabular}{@{}l c c@{}}
\toprule
\textbf{Method} & $\Delta$Accuracy (pp) & $\Delta$Alignment Score \\
\midrule
ACSS-TS   & $-0.41$ & $-0.34$ \\
ACSS-RR   & $-0.88$ & $+0.67$ \\
ACSS-Ali. & $+0.57$ & $+7.99$ \\
SC-Align  & $+0.38$ & $+7.66$ \\
\bottomrule
\end{tabular}
\caption{Average improvements of alignment-aware selection methods relative to SC-CoT across all models.}
\label{tab:avg_delta_vs_sccot}
\end{table}

\paragraph{Correlation with accuracy.}
Table~\ref{tab:cross_model_grouped} reveals three consistent empirical observations: (i) Regarding ACSS-TS and ACSS-RR, which select chains by minimizing specific alignment errors, these methods clearly improve both Accuracy and Alignment Score over the CoT baseline in most cases, but usually do not beat SC-CoT in Accuracy;
(ii) when we use the Alignment Score itself as the selection signal (ACSS-Ali.), several models, including GPT-4o-mini, GPT-4o, and Qwen2.5-3B, achieve higher accuracy than SC-CoT while also enjoying substantial gains in Alignment Score;
(iii) this effect is pronounced for reasoning models, where base CoT performance is strong, so SC-CoT yields diminishing or negative gains, matching studies where CoT adds little on strong models~\citep{wang-etal-2025-ranked,liu2025mindstepbystep}, whereas ACSS-Ali. still surpasses SC-CoT in accuracy.
To make this coupling explicit, Table~\ref{tab:avg_delta_vs_sccot} summarizes the average improvements over SC-CoT: alignment-aware methods that directly optimize Alignment Score (ACSS-Ali.\ and SC-Align) deliver the largest gains in Alignment Score while improving Accuracy. \uline{These observations indicate an empirical coupling between Alignment Score and answer correctness.}

\paragraph{High-accuracy chains are more aligned.}
To test whether high-accuracy chains are \emph{intrinsically} more aligned, we further examine SC-Align, which first performs majority voting as in SC-CoT and then resolves ties by selecting the chain with the highest Alignment Score. Consistent with our theoretical view, Table~\ref{tab:cross_model_grouped} shows that SC-Align closely matches SC-CoT in Accuracy across all models, while achieving markedly higher Alignment Scores, often by a large margin. \uline{This pattern suggests that chains with higher probability of being correct naturally exhibit higher semantic alignment.}

\paragraph{LLM-based evaluation with ablations.}
Table~\ref{tab:scalign_judge_vertical} reports LLM-based evaluation for SC-CoT and SC-Align under three distinct NLI backbones~\cite{liu2019roberta,he2023debertav,lewis-etal-2020-bart}.  
Across all NLI backbones, SC-Align attains higher mean scores on both coherence and readability and is preferred more frequently in pairwise comparisons. We additionally conduct an extensive grid ablation (see Appendix~\ref{subsec:judge-ablation}) and observe the same pattern.
These consistent trends indicate that: \uline{Alignment Score is insensitive to the specific NLI backbone and behaves as a stable and reliable signal that tracks human preferences.} We also provide a qualitative case study about the evaluation results, which can be found in Appendix~\ref{app:qualitative-cases}.

\begin{table}[t]
\centering
\small  
\setlength{\tabcolsep}{5pt}   
\renewcommand{\arraystretch}{0.9}  
\begin{tabular}{@{}p{1.3cm} l c c@{}}
\toprule
\textbf{Aspect} & \textbf{Method} & \textbf{Score (mean$\pm$std)} & \textbf{Pairwise} \\
\midrule

\rowcolor{gray!15}
\multicolumn{4}{c}{\textit{NLI Model: roberta-large-mnli}} \\
\cmidrule(lr){1-4}
\multirow{2}{*}{Coherence} 
  & SC-CoT   & 8.54{\scriptsize($\pm 0.78$)} & 12.0\% win \\
  & SC-Align & \textbf{8.69}{\scriptsize($\pm 0.78$)} & \textbf{28.5\% win} \\
\midrule
\multirow{2}{*}{Readability} 
  & SC-CoT   & 8.23{\scriptsize($\pm 0.58$)} & 22.5\% win \\
  & SC-Align & \textbf{8.45}{\scriptsize($\pm 0.59$)} & \textbf{45.5\% win} \\

\midrule
\rowcolor{gray!15}
\multicolumn{4}{c}{\textit{NLI Model: deberta-v3-large-mnli}} \\
\cmidrule(lr){1-4}
\multirow{2}{*}{Coherence} 
  & SC-CoT   & 8.49{\scriptsize($\pm 0.84$)} & 20.0\% win \\
  & SC-Align & \textbf{8.54}{\scriptsize($\pm 0.92$)} & \textbf{28.5\% win} \\
\midrule
\multirow{2}{*}{Readability} 
  & SC-CoT   & 8.30{\scriptsize($\pm 0.61$)} & 26.5\% win \\
  & SC-Align & \textbf{8.44}{\scriptsize($\pm 0.64$)} & \textbf{43.0\% win} \\

\midrule
\rowcolor{gray!15}
\multicolumn{4}{c}{\textit{NLI Model: bart-large-mnli}} \\
\cmidrule(lr){1-4}
\multirow{2}{*}{Coherence} 
  & SC-CoT   & 8.55{\scriptsize($\pm 0.84$)} & 14.0\% win \\
  & SC-Align & \textbf{8.63}{\scriptsize($\pm 0.95$)} & \textbf{26.5\% win} \\
\midrule
\multirow{2}{*}{Readability} 
  & SC-CoT   & 8.22{\scriptsize($\pm 0.59$)} & 19.5\% win \\
  & SC-Align & \textbf{8.50}{\scriptsize($\pm 0.61$)} & \textbf{49.0\% win} \\
\bottomrule
\end{tabular}
\caption{LLM-based evaluation of reasoning quality (SC-CoT vs.\ SC-Align) under three distinct NLI backbones. Scores are on a 1--10 scale.}
\label{tab:scalign_judge_vertical}
\end{table}

\section{Discussion: Applicability Beyond CoT}
\label{sec:discussion_beyond_cot}

Although we study Alignment Score under standard CoT prompting, the metric is not tied to CoT itself. Rather, CoT serves as a controlled setting in which intermediate reasoning states are explicitly segmented into ordered steps. More generally, Alignment Score applies to structured reasoning trajectories whose intermediate states can be decomposed into comparable units and analyzed through their pairwise semantic relations. This broader view is also motivated by recent work showing that intermediate trajectories in reasoning and agent-based systems may themselves be vulnerable to failure and manipulation~\cite{Hu_Dong_Sun_Huang_2026,hu2026lyingtruthsopenchannelmultiagent}.

\paragraph{Generalizability to other reasoning frameworks.}
Here, we discuss the applicability of the Alignment Score to representative reasoning frameworks. In Reflexion~\cite{shinn2023reflexion}, the initial response, critique, and revised response can be treated as staged segments. In ReAct~\cite{yao2023react}, each thought--action-observation cycle can be viewed as a structured step, allowing consistency to be assessed across reasoning and interaction. More broadly, the framework may apply to agent-style reasoning settings that use explicit intermediate trajectories, although more complex situations, such as branching or partially ordered reasoning, would need further alignment and segmentation design.



\section{Conclusion}
\label{sec:conclusion}
In this paper, we take CoT as a concrete entry point for measuring \emph{structured reasoning alignment} in large language models. We introduce the \textit{Alignment Score}, a lightweight measure that aligns generated chains with reference demonstrations in a one-to-one manner, and show across diverse settings that it is coupled with task accuracy and overall quality. As CoT demonstrations (reference chains) become more logically coherent, readable, correct, and better aligned with human preferences, the Alignment Score increases and accuracy improves. Moreover, we empirically show that optimizing the Alignment Score will lead to increased readability and logical coherence.  In this sense, CoT moves from being a prompting heuristic to serving as a quantitative lens on reasoning quality and a diagnostic signal.

\section*{Limitations}

This study is subject to several limitations:
\begin{itemize}[leftmargin=*,topsep=2pt,itemsep=2pt,parsep=0pt]
    \item The proposed Alignment Score is designed for \emph{step-structured} reasoning traces with clearly separated statements. Many real-world explanations are written as free-form paragraphs without explicit step boundaries and therefore require additional processing. In such settings, our entropy-based alignment framework is not directly applicable. Extending Alignment Score to less structured forms of reasoning remains a challenge.
    \item Our experiments are conducted on multiple-choice QA benchmarks with a single correct option. These settings provide a clean testbed, but they do not cover open-ended tasks such as code generation. How well Alignment Score and our alignment-aware methods generalize to these broader scenarios remains to be tested.
    \item In Section~\ref{sec:rq2_peak}, we empirically show that reasoning chains with more hops are more susceptible to certain alignment errors. However, we do not yet provide a theoretical account or formal characterization of why this degradation emerges.
\end{itemize}

\section*{Ethical Considerations}
Our work is primarily methodological, but it still raises several ethical considerations about evaluation bias, interpretation, and potential deployment.
\paragraph{Data and privacy.}
This work uses public multiple-choice QA benchmarks and AI-assisted reference-chain generation, and therefore does not involve personally identifiable information. We checked data and case reporting to ensure that no information requiring anonymization is included.

\paragraph{Potential bias in LLM-based evaluation.}
Because Alignment Score is defined relative to curated reference chains, it may prefer some reasoning styles over other equally valid ones. In addition, our LLM-as-a-judge pipeline may introduce stylistic bias when evaluating readability and coherence.

\paragraph{Interpretation and deployment.}
A high Alignment Score must not be taken as a guarantee of truthfulness, accuracy, or safety. We regard the Alignment Score as a diagnostic indicator, not as a sufficient condition on its own for use in high-risk domains, including medical applications.

\section*{Acknowledgments}
This work is partially funded by the European Union (under grant agreement ID 101212818). Views and opinions expressed are however those of the author(s) only and do not necessarily reflect those of the European Union or European Health and Digital Executive Agency (HADEA). Neither the European Union nor the granting authority can be held responsible for them. 
This work is partially supported by Innovate UK through AI-PASSPORT under Grant 10126404. This work was awarded a grant by the AI Security Institute (AISI) via the Alignment Project
(Rare-Event Estimation in Large Language Models via Subset Simulation) and funded by EPSRC.
Yi's contribution is partially supported through the Royal Society international exchanges programme and in part by the Engineering and Physical Sciences Research Council, through funding from RAi UK [EP/Y009800/1].

\bibliography{custom}

@inproceedings{lampinen2022can,
    title = "Can language models learn from explanations in context?",
    author = "Lampinen, Andrew  and
      Dasgupta, Ishita  and
      Chan, Stephanie  and
      Mathewson, Kory  and
      Tessler, Mh  and
      Creswell, Antonia  and
      McClelland, James  and
      Wang, Jane  and
      Hill, Felix",
    editor = "Goldberg, Yoav  and
      Kozareva, Zornitsa  and
      Zhang, Yue",
    booktitle = "Findings of the Association for Computational Linguistics: EMNLP 2022",
    month = dec,
    year = "2022",
    address = "Abu Dhabi, United Arab Emirates",
    publisher = "Association for Computational Linguistics",
    url = "https://aclanthology.org/2022.findings-emnlp.38/",
    doi = "10.18653/v1/2022.findings-emnlp.38",
    pages = "537--563"
}

@inproceedings{dong2022survey,
    title = "A Survey on In-context Learning",
    author = "Dong, Qingxiu  and
      Li, Lei  and
      Dai, Damai  and
      Zheng, Ce  and
      Ma, Jingyuan  and
      Li, Rui  and
      Xia, Heming  and
      Xu, Jingjing  and
      Wu, Zhiyong  and
      Chang, Baobao  and
      Sun, Xu  and
      Li, Lei  and
      Sui, Zhifang",
    editor = "Al-Onaizan, Yaser  and
      Bansal, Mohit  and
      Chen, Yun-Nung",
    booktitle = "Proceedings of the 2024 Conference on Empirical Methods in Natural Language Processing",
    month = nov,
    year = "2024",
    address = "Miami, Florida, USA",
    publisher = "Association for Computational Linguistics",
    url = "https://aclanthology.org/2024.emnlp-main.64/",
    doi = "10.18653/v1/2024.emnlp-main.64",
    pages = "1107--1128"
}

@inproceedings{wei2022cot,
 author = {Wei, Jason and Wang, Xuezhi and Schuurmans, Dale and Bosma, Maarten and ichter, brian and Xia, Fei and Chi, Ed and Le, Quoc V and Zhou, Denny},
 booktitle = {Advances in Neural Information Processing Systems},
 editor = {S. Koyejo and S. Mohamed and A. Agarwal and D. Belgrave and K. Cho and A. Oh},
 pages = {24824--24837},
 publisher = {Curran Associates, Inc.},
 title = {Chain-of-Thought Prompting Elicits Reasoning in Large Language Models},
 url = {https://proceedings.neurips.cc/paper_files/paper/2022/file/9d5609613524ecf4f15af0f7b31abca4-Paper-Conference.pdf},
 volume = {35},
 year = {2022}
}

@inproceedings{nguyen2024direct,
    title = "Direct Evaluation of Chain-of-Thought in Multi-hop Reasoning with Knowledge Graphs",
    author = "Nguyen, Minh-Vuong  and
      Luo, Linhao  and
      Shiri, Fatemeh  and
      Phung, Dinh  and
      Li, Yuan-Fang  and
      Vu, Thuy-Trang  and
      Haffari, Gholamreza",
    editor = "Ku, Lun-Wei  and
      Martins, Andre  and
      Srikumar, Vivek",
    booktitle = "Findings of the Association for Computational Linguistics: ACL 2024",
    month = aug,
    year = "2024",
    address = "Bangkok, Thailand",
    publisher = "Association for Computational Linguistics",
    url = "https://aclanthology.org/2024.findings-acl.168/",
    doi = "10.18653/v1/2024.findings-acl.168",
    pages = "2862--2883"
}

@inproceedings{lu2022learn,
 author = {Lu, Pan and Mishra, Swaroop and Xia, Tanglin and Qiu, Liang and Chang, Kai-Wei and Zhu, Song-Chun and Tafjord, Oyvind and Clark, Peter and Kalyan, Ashwin},
 booktitle = {Advances in Neural Information Processing Systems},
 editor = {S. Koyejo and S. Mohamed and A. Agarwal and D. Belgrave and K. Cho and A. Oh},
 pages = {2507--2521},
 publisher = {Curran Associates, Inc.},
 title = {Learn to Explain: Multimodal Reasoning via Thought Chains for Science Question Answering},
 url = {https://proceedings.neurips.cc/paper_files/paper/2022/file/11332b6b6cf4485b84afadb1352d3a9a-Paper-Conference.pdf},
 volume = {35},
 year = {2022}
}

@misc{chollet2019measure,
      title={On the Measure of Intelligence}, 
      author={François Chollet},
      year={2019},
      eprint={1911.01547},
      archivePrefix={arXiv},
      primaryClass={cs.AI},
      url={https://arxiv.org/abs/1911.01547}, 
}

@inproceedings{kojima2023largelanguagemodelszeroshot,
 author = {Kojima, Takeshi and Gu, Shixiang (Shane) and Reid, Machel and Matsuo, Yutaka and Iwasawa, Yusuke},
 booktitle = {Advances in Neural Information Processing Systems},
 editor = {S. Koyejo and S. Mohamed and A. Agarwal and D. Belgrave and K. Cho and A. Oh},
 pages = {22199--22213},
 publisher = {Curran Associates, Inc.},
 title = {Large Language Models are Zero-Shot Reasoners},
 url = {https://proceedings.neurips.cc/paper_files/paper/2022/file/8bb0d291acd4acf06ef112099c16f326-Paper-Conference.pdf},
 volume = {35},
 year = {2022}
}

@inproceedings{yang2018hotpotqa,
    title = "{H}otpot{QA}: A Dataset for Diverse, Explainable Multi-hop Question Answering",
    author = "Yang, Zhilin  and
      Qi, Peng  and
      Zhang, Saizheng  and
      Bengio, Yoshua  and
      Cohen, William  and
      Salakhutdinov, Ruslan  and
      Manning, Christopher D.",
    editor = "Riloff, Ellen  and
      Chiang, David  and
      Hockenmaier, Julia  and
      Tsujii, Jun{'}ichi",
    booktitle = "Proceedings of the 2018 Conference on Empirical Methods in Natural Language Processing",
    month = oct # "-" # nov,
    year = "2018",
    address = "Brussels, Belgium",
    publisher = "Association for Computational Linguistics",
    url = "https://aclanthology.org/D18-1259/",
    doi = "10.18653/v1/D18-1259",
    pages = "2369--2380"
}

@inproceedings{lin2022truthfulqa,
    title = "{T}ruthful{QA}: Measuring How Models Mimic Human Falsehoods",
    author = "Lin, Stephanie  and
      Hilton, Jacob  and
      Evans, Owain",
    editor = "Muresan, Smaranda  and
      Nakov, Preslav  and
      Villavicencio, Aline",
    booktitle = "Proceedings of the 60th Annual Meeting of the Association for Computational Linguistics (Volume 1: Long Papers)",
    month = may,
    year = "2022",
    address = "Dublin, Ireland",
    publisher = "Association for Computational Linguistics",
    url = "https://aclanthology.org/2022.acl-long.229/",
    doi = "10.18653/v1/2022.acl-long.229",
    pages = "3214--3252"
}

@inproceedings{
zhou2022least,
title={Least-to-Most Prompting Enables Complex Reasoning in Large Language Models},
author={Denny Zhou and Nathanael Sch{\"a}rli and Le Hou and Jason Wei and Nathan Scales and Xuezhi Wang and Dale Schuurmans and Claire Cui and Olivier Bousquet and Quoc V Le and Ed H. Chi},
booktitle={The Eleventh International Conference on Learning Representations },
year={2023},
url={https://openreview.net/forum?id=WZH7099tgfM}
}

@inproceedings{li2016diversity,
    title = "A Diversity-Promoting Objective Function for Neural Conversation Models",
    author = "Li, Jiwei  and
      Galley, Michel  and
      Brockett, Chris  and
      Gao, Jianfeng  and
      Dolan, Bill",
    editor = "Knight, Kevin  and
      Nenkova, Ani  and
      Rambow, Owen",
    booktitle = "Proceedings of the 2016 Conference of the North {A}merican Chapter of the Association for Computational Linguistics: Human Language Technologies",
    month = jun,
    year = "2016",
    address = "San Diego, California",
    publisher = "Association for Computational Linguistics",
    url = "https://aclanthology.org/N16-1014/",
    doi = "10.18653/v1/N16-1014",
    pages = "110--119"
}

@ARTICLE{lin1991divergence,
  author={Lin, J.},
  journal={IEEE Transactions on Information Theory}, 
  title={Divergence measures based on the Shannon entropy}, 
  year={1991},
  volume={37},
  number={1},
  pages={145-151},
  doi={10.1109/18.61115}}

@article{
srivastava2022bigbench,
title={Beyond the Imitation Game: Quantifying and extrapolating the capabilities of language models},
author={Aarohi Srivastava and Abhinav Rastogi and Abhishek Rao and Abu Awal Md Shoeb and Abubakar Abid and Adam Fisch and Adam R. Brown and Adam Santoro and Aditya Gupta and Adri{\`a} Garriga-Alonso and Agnieszka Kluska and Aitor Lewkowycz and Akshat Agarwal and Alethea Power and Alex Ray and Alex Warstadt and Alexander W. Kocurek and Ali Safaya and Ali Tazarv and Alice Xiang and Alicia Parrish and Allen Nie and Aman Hussain and Amanda Askell and Amanda Dsouza and Ambrose Slone and Ameet Rahane and Anantharaman S. Iyer and Anders Johan Andreassen and Andrea Madotto and Andrea Santilli and Andreas Stuhlm{\"u}ller and Andrew M. Dai and Andrew La and Andrew Kyle Lampinen and Andy Zou and Angela Jiang and Angelica Chen and Anh Vuong and Animesh Gupta and Anna Gottardi and Antonio Norelli and Anu Venkatesh and Arash Gholamidavoodi and Arfa Tabassum and Arul Menezes and Arun Kirubarajan and Asher Mullokandov and Ashish Sabharwal and Austin Herrick and Avia Efrat and Aykut Erdem and Ayla Karaka{\c{s}} and B. Ryan Roberts and Bao Sheng Loe and Barret Zoph and Bart{\l}omiej Bojanowski and Batuhan {\"O}zyurt and Behnam Hedayatnia and Behnam Neyshabur and Benjamin Inden and Benno Stein and Berk Ekmekci and Bill Yuchen Lin and Blake Howald and Bryan Orinion and Cameron Diao and Cameron Dour and Catherine Stinson and Cedrick Argueta and Cesar Ferri and Chandan Singh and Charles Rathkopf and Chenlin Meng and Chitta Baral and Chiyu Wu and Chris Callison-Burch and Christopher Waites and Christian Voigt and Christopher D Manning and Christopher Potts and Cindy Ramirez and Clara E. Rivera and Clemencia Siro and Colin Raffel and Courtney Ashcraft and Cristina Garbacea and Damien Sileo and Dan Garrette and Dan Hendrycks and Dan Kilman and Dan Roth and C. Daniel Freeman and Daniel Khashabi and Daniel Levy and Daniel Mosegu{\'\i} Gonz{\'a}lez and Danielle Perszyk and Danny Hernandez and Danqi Chen and Daphne Ippolito and Dar Gilboa and David Dohan and David Drakard and David Jurgens and Debajyoti Datta and Deep Ganguli and Denis Emelin and Denis Kleyko and Deniz Yuret and Derek Chen and Derek Tam and Dieuwke Hupkes and Diganta Misra and Dilyar Buzan and Dimitri Coelho Mollo and Diyi Yang and Dong-Ho Lee and Dylan Schrader and Ekaterina Shutova and Ekin Dogus Cubuk and Elad Segal and Eleanor Hagerman and Elizabeth Barnes and Elizabeth Donoway and Ellie Pavlick and Emanuele Rodol{\`a} and Emma Lam and Eric Chu and Eric Tang and Erkut Erdem and Ernie Chang and Ethan A Chi and Ethan Dyer and Ethan Jerzak and Ethan Kim and Eunice Engefu Manyasi and Evgenii Zheltonozhskii and Fanyue Xia and Fatemeh Siar and Fernando Mart{\'\i}nez-Plumed and Francesca Happ{\'e} and Francois Chollet and Frieda Rong and Gaurav Mishra and Genta Indra Winata and Gerard de Melo and Germ{\`a}n Kruszewski and Giambattista Parascandolo and Giorgio Mariani and Gloria Xinyue Wang and Gonzalo Jaimovitch-Lopez and Gregor Betz and Guy Gur-Ari and Hana Galijasevic and Hannah Kim and Hannah Rashkin and Hannaneh Hajishirzi and Harsh Mehta and Hayden Bogar and Henry Francis Anthony Shevlin and Hinrich Schuetze and Hiromu Yakura and Hongming Zhang and Hugh Mee Wong and Ian Ng and Isaac Noble and Jaap Jumelet and Jack Geissinger and Jackson Kernion and Jacob Hilton and Jaehoon Lee and Jaime Fern{\'a}ndez Fisac and James B Simon and James Koppel and James Zheng and James Zou and Jan Kocon and Jana Thompson and Janelle Wingfield and Jared Kaplan and Jarema Radom and Jascha Sohl-Dickstein and Jason Phang and Jason Wei and Jason Yosinski and Jekaterina Novikova and Jelle Bosscher and Jennifer Marsh and Jeremy Kim and Jeroen Taal and Jesse Engel and Jesujoba Alabi and Jiacheng Xu and Jiaming Song and Jillian Tang and Joan Waweru and John Burden and John Miller and John U. Balis and Jonathan Batchelder and Jonathan Berant and J{\"o}rg Frohberg and Jos Rozen and Jose Hernandez-Orallo and Joseph Boudeman and Joseph Guerr and Joseph Jones and Joshua B. Tenenbaum and Joshua S. Rule and Joyce Chua and Kamil Kanclerz and Karen Livescu and Karl Krauth and Karthik Gopalakrishnan and Katerina Ignatyeva and Katja Markert and Kaustubh Dhole and Kevin Gimpel and Kevin Omondi and Kory Wallace Mathewson and Kristen Chiafullo and Ksenia Shkaruta and Kumar Shridhar and Kyle McDonell and Kyle Richardson and Laria Reynolds and Leo Gao and Li Zhang and Liam Dugan and Lianhui Qin and Lidia Contreras-Ochando and Louis-Philippe Morency and Luca Moschella and Lucas Lam and Lucy Noble and Ludwig Schmidt and Luheng He and Luis Oliveros-Col{\'o}n and Luke Metz and L{\"u}tfi Kerem Senel and Maarten Bosma and Maarten Sap and Maartje Ter Hoeve and Maheen Farooqi and Manaal Faruqui and Mantas Mazeika and Marco Baturan and Marco Marelli and Marco Maru and Maria Jose Ramirez-Quintana and Marie Tolkiehn and Mario Giulianelli and Martha Lewis and Martin Potthast and Matthew L Leavitt and Matthias Hagen and M{\'a}ty{\'a}s Schubert and Medina Orduna Baitemirova and Melody Arnaud and Melvin McElrath and Michael Andrew Yee and Michael Cohen and Michael Gu and Michael Ivanitskiy and Michael Starritt and Michael Strube and Micha{\l} Sw{\k{e}}drowski and Michele Bevilacqua and Michihiro Yasunaga and Mihir Kale and Mike Cain and Mimee Xu and Mirac Suzgun and Mitch Walker and Mo Tiwari and Mohit Bansal and Moin Aminnaseri and Mor Geva and Mozhdeh Gheini and Mukund Varma T and Nanyun Peng and Nathan Andrew Chi and Nayeon Lee and Neta Gur-Ari Krakover and Nicholas Cameron and Nicholas Roberts and Nick Doiron and Nicole Martinez and Nikita Nangia and Niklas Deckers and Niklas Muennighoff and Nitish Shirish Keskar and Niveditha S. Iyer and Noah Constant and Noah Fiedel and Nuan Wen and Oliver Zhang and Omar Agha and Omar Elbaghdadi and Omer Levy and Owain Evans and Pablo Antonio Moreno Casares and Parth Doshi and Pascale Fung and Paul Pu Liang and Paul Vicol and Pegah Alipoormolabashi and Peiyuan Liao and Percy Liang and Peter W Chang and Peter Eckersley and Phu Mon Htut and Pinyu Hwang and Piotr Mi{\l}kowski and Piyush Patil and Pouya Pezeshkpour and Priti Oli and Qiaozhu Mei and Qing Lyu and Qinlang Chen and Rabin Banjade and Rachel Etta Rudolph and Raefer Gabriel and Rahel Habacker and Ramon Risco and Rapha{\"e}l Milli{\`e}re and Rhythm Garg and Richard Barnes and Rif A. Saurous and Riku Arakawa and Robbe Raymaekers and Robert Frank and Rohan Sikand and Roman Novak and Roman Sitelew and Ronan Le Bras and Rosanne Liu and Rowan Jacobs and Rui Zhang and Russ Salakhutdinov and Ryan Andrew Chi and Seungjae Ryan Lee and Ryan Stovall and Ryan Teehan and Rylan Yang and Sahib Singh and Saif M. Mohammad and Sajant Anand and Sam Dillavou and Sam Shleifer and Sam Wiseman and Samuel Gruetter and Samuel R. Bowman and Samuel Stern Schoenholz and Sanghyun Han and Sanjeev Kwatra and Sarah A. Rous and Sarik Ghazarian and Sayan Ghosh and Sean Casey and Sebastian Bischoff and Sebastian Gehrmann and Sebastian Schuster and Sepideh Sadeghi and Shadi Hamdan and Sharon Zhou and Shashank Srivastava and Sherry Shi and Shikhar Singh and Shima Asaadi and Shixiang Shane Gu and Shubh Pachchigar and Shubham Toshniwal and Shyam Upadhyay and Shyamolima Shammie Debnath and Siamak Shakeri and Simon Thormeyer and Simone Melzi and Siva Reddy and Sneha Priscilla Makini and Soo-Hwan Lee and Spencer Torene and Sriharsha Hatwar and Stanislas Dehaene and Stefan Divic and Stefano Ermon and Stella Biderman and Stephanie Lin and Stephen Prasad and Steven Piantadosi and Stuart Shieber and Summer Misherghi and Svetlana Kiritchenko and Swaroop Mishra and Tal Linzen and Tal Schuster and Tao Li and Tao Yu and Tariq Ali and Tatsunori Hashimoto and Te-Lin Wu and Th{\'e}o Desbordes and Theodore Rothschild and Thomas Phan and Tianle Wang and Tiberius Nkinyili and Timo Schick and Timofei Kornev and Titus Tunduny and Tobias Gerstenberg and Trenton Chang and Trishala Neeraj and Tushar Khot and Tyler Shultz and Uri Shaham and Vedant Misra and Vera Demberg and Victoria Nyamai and Vikas Raunak and Vinay Venkatesh Ramasesh and vinay uday prabhu and Vishakh Padmakumar and Vivek Srikumar and William Fedus and William Saunders and William Zhang and Wout Vossen and Xiang Ren and Xiaoyu Tong and Xinran Zhao and Xinyi Wu and Xudong Shen and Yadollah Yaghoobzadeh and Yair Lakretz and Yangqiu Song and Yasaman Bahri and Yejin Choi and Yichi Yang and Sophie Hao and Yifu Chen and Yonatan Belinkov and Yu Hou and Yufang Hou and Yuntao Bai and Zachary Seid and Zhuoye Zhao and Zijian Wang and Zijie J. Wang and Zirui Wang and Ziyi Wu},
journal={Transactions on Machine Learning Research},
issn={2835-8856},
year={2023},
url={https://openreview.net/forum?id=uyTL5Bvosj},
note={Featured Certification}
}

@inproceedings{jacovi2024reveal,
    title = "A Chain-of-Thought Is as Strong as Its Weakest Link: A Benchmark for Verifiers of Reasoning Chains",
    author = "Jacovi, Alon  and
      Bitton, Yonatan  and
      Bohnet, Bernd  and
      Herzig, Jonathan  and
      Honovich, Or  and
      Tseng, Michael  and
      Collins, Michael  and
      Aharoni, Roee  and
      Geva, Mor",
    editor = "Ku, Lun-Wei  and
      Martins, Andre  and
      Srikumar, Vivek",
    booktitle = "Proceedings of the 62nd Annual Meeting of the Association for Computational Linguistics (Volume 1: Long Papers)",
    month = aug,
    year = "2024",
    address = "Bangkok, Thailand",
    publisher = "Association for Computational Linguistics",
    url = "https://aclanthology.org/2024.acl-long.254/",
    doi = "10.18653/v1/2024.acl-long.254",
    pages = "4615--4634"
}

@misc{vacareanu2024general,
  title         = {General Purpose Verification for Chain of Thought Prompting},
  author        = {Robert Vacareanu and Anurag Pratik and Evangelia Spiliopoulou and Zheng Qi and Giovanni Paolini and Neha Anna John and Jie Ma and Yassine Benajiba and Miguel Ballesteros},
  year          = {2024},
  eprint        = {2405.00204},
  archivePrefix = {arXiv},
  primaryClass  = {cs.CL}
}

@inproceedings{
wang2022selfconsistency,
title={Self-Consistency Improves Chain of Thought Reasoning in Language Models},
author={Xuezhi Wang and Jason Wei and Dale Schuurmans and Quoc V Le and Ed H. Chi and Sharan Narang and Aakanksha Chowdhery and Denny Zhou},
booktitle={The Eleventh International Conference on Learning Representations },
year={2023},
url={https://openreview.net/forum?id=1PL1NIMMrw}
}

@inproceedings{suzgun2022challenging,
    title = "Challenging {BIG}-Bench Tasks and Whether Chain-of-Thought Can Solve Them",
    author = {Suzgun, Mirac  and
      Scales, Nathan  and
      Sch{\"a}rli, Nathanael  and
      Gehrmann, Sebastian  and
      Tay, Yi  and
      Chung, Hyung Won  and
      Chowdhery, Aakanksha  and
      Le, Quoc  and
      Chi, Ed  and
      Zhou, Denny  and
      Wei, Jason},
    editor = "Rogers, Anna  and
      Boyd-Graber, Jordan  and
      Okazaki, Naoaki",
    booktitle = "Findings of the Association for Computational Linguistics: ACL 2023",
    month = jul,
    year = "2023",
    address = "Toronto, Canada",
    publisher = "Association for Computational Linguistics",
    url = "https://aclanthology.org/2023.findings-acl.824/",
    doi = "10.18653/v1/2023.findings-acl.824",
    pages = "13003--13051"
}

@article{cowan2001magical,
  title={The magical number 4 in short-term memory: A reconsideration of mental storage capacity},
  author={Cowan, Nelson},
  journal={Behavioral and Brain Sciences},
  volume={24},
  number={1},
  pages={87--114},
  year={2001},
  publisher={Cambridge University Press},
  doi={10.1017/S0140525X01003922}
}

@misc{cobbe2021gsm8k,
      title={Training Verifiers to Solve Math Word Problems}, 
      author={Karl Cobbe and Vineet Kosaraju and Mohammad Bavarian and Mark Chen and Heewoo Jun and Lukasz Kaiser and Matthias Plappert and Jerry Tworek and Jacob Hilton and Reiichiro Nakano and Christopher Hesse and John Schulman},
      year={2021},
      eprint={2110.14168},
      archivePrefix={arXiv},
      primaryClass={cs.LG},
      url={https://arxiv.org/abs/2110.14168}, 
}

@inproceedings{
turpin2023language,
title={Language Models Don't Always Say What They Think: Unfaithful Explanations in Chain-of-Thought Prompting},
author={Miles Turpin and Julian Michael and Ethan Perez and Samuel R. Bowman},
booktitle={Thirty-seventh Conference on Neural Information Processing Systems},
year={2023},
url={https://openreview.net/forum?id=bzs4uPLXvi}
}

@inproceedings{
kuhn2023semantic,
title={Semantic Uncertainty: Linguistic Invariances for Uncertainty Estimation in Natural Language Generation},
author={Lorenz Kuhn and Yarin Gal and Sebastian Farquhar},
booktitle={The Eleventh International Conference on Learning Representations },
year={2023},
url={https://openreview.net/forum?id=VD-AYtP0dve}
}

@inproceedings{
rein2024gpqa,
title={{GPQA}: A Graduate-Level Google-Proof Q\&A Benchmark},
author={David Rein and Betty Li Hou and Asa Cooper Stickland and Jackson Petty and Richard Yuanzhe Pang and Julien Dirani and Julian Michael and Samuel R. Bowman},
booktitle={First Conference on Language Modeling},
year={2024},
url={https://openreview.net/forum?id=Ti67584b98}
}

@inproceedings{
hendrycks2021m,
title={Measuring Massive Multitask Language Understanding},
author={Dan Hendrycks and Collin Burns and Steven Basart and Andy Zou and Mantas Mazeika and Dawn Song and Jacob Steinhardt},
booktitle={International Conference on Learning Representations},
year={2021},
url={https://openreview.net/forum?id=d7KBjmI3GmQ}
}

@article{farquhar2024detecting,
  title={Detecting hallucinations in large language models using semantic entropy},
  author={Farquhar, Sebastian and Kossen, Jannik and Kuhn, Lorenz and Gal, Yarin},
  journal={Nature},
  volume={630},
  number={8017},
  pages={625--630},
  year={2024},
  publisher={Nature Publishing Group UK London}
}

@misc{zhu2025shapley,
      title={Shapley Uncertainty in Natural Language Generation}, 
      author={Meilin Zhu and Gaojie Jin and Xiaowei Huang and Lijun Zhang},
      year={2025},
      eprint={2507.21406},
      archivePrefix={arXiv},
      primaryClass={cs.AI},
      url={https://arxiv.org/abs/2507.21406}, 
}

@inproceedings{nikitin2024kernel,
 author = {Nikitin, Alexander and Kossen, Jannik and Gal, Yarin and Marttinen, Pekka},
 booktitle = {Advances in Neural Information Processing Systems},
 doi = {10.52202/079017-0283},
 editor = {A. Globerson and L. Mackey and D. Belgrave and A. Fan and U. Paquet and J. Tomczak and C. Zhang},
 pages = {8901--8929},
 publisher = {Curran Associates, Inc.},
 title = {Kernel Language Entropy: Fine-grained Uncertainty Quantification for LLMs from Semantic Similarities},
 url = {https://proceedings.neurips.cc/paper_files/paper/2024/file/10c456d2160517581a234dfde15a7505-Paper-Conference.pdf},
 volume = {37},
 year = {2024}
}

@misc{ma2025semantic,
      title={Semantic Energy: Detecting LLM Hallucination Beyond Entropy}, 
      author={Huan Ma and Jiadong Pan and Jing Liu and Yan Chen and Joey Tianyi Zhou and Guangyu Wang and Qinghua Hu and Hua Wu and Changqing Zhang and Haifeng Wang},
      year={2025},
      eprint={2508.14496},
      archivePrefix={arXiv},
      primaryClass={cs.LG},
      url={https://arxiv.org/abs/2508.14496}, 
}

@misc{Clark2018ThinkYH,
      title={Think you have Solved Question Answering? Try ARC, the AI2 Reasoning Challenge}, 
      author={Peter Clark and Isaac Cowhey and Oren Etzioni and Tushar Khot and Ashish Sabharwal and Carissa Schoenick and Oyvind Tafjord},
      year={2018},
      eprint={1803.05457},
      archivePrefix={arXiv},
      primaryClass={cs.AI},
      url={https://arxiv.org/abs/1803.05457}, 
}

@misc{tii2023falconllm,
  author       = {{Technology Innovation Institute}},
  title        = {{Falcon-7B-Instruct}},
  year         = {2023},
  howpublished = {Hugging Face Model Card},
  url          = {https://huggingface.co/tiiuae/falcon-7b-instruct},
  note         = {Accessed: 2025-12-26}
}

@misc{qwen2_5_2025,
      title={Qwen2.5 Technical Report}, 
      author={An Yang and Baosong Yang and Beichen Zhang and Binyuan Hui and Bo Zheng and Bowen Yu and Chengyuan Li and Dayiheng Liu and Fei Huang and Haoran Wei and Huan Lin and Jian Yang and Jianhong Tu and Jianwei Zhang and Jianxin Yang and Jiaxi Yang and Jingren Zhou and Junyang Lin and Kai Dang and Keming Lu and Keqin Bao and Kexin Yang and Le Yu and Mei Li and Mingfeng Xue and Pei Zhang and Qin Zhu and Rui Men and Runji Lin and Tianhao Li and Tianyi Tang and Tingyu Xia and Xingzhang Ren and Xuancheng Ren and Yang Fan and Yang Su and Yichang Zhang and Yu Wan and Yuqiong Liu and Zeyu Cui and Zhenru Zhang and Zihan Qiu},
      year={2025},
      eprint={2412.15115},
      archivePrefix={arXiv},
      primaryClass={cs.CL},
      url={https://arxiv.org/abs/2412.15115}, 
}

@misc{touvron2023llama,
      title={Llama 2: Open Foundation and Fine-Tuned Chat Models}, 
      author={Hugo Touvron and Louis Martin and Kevin Stone and Peter Albert and Amjad Almahairi and Yasmine Babaei and Nikolay Bashlykov and Soumya Batra and Prajjwal Bhargava and Shruti Bhosale and Dan Bikel and Lukas Blecher and Cristian Canton Ferrer and Moya Chen and Guillem Cucurull and David Esiobu and Jude Fernandes and Jeremy Fu and Wenyin Fu and Brian Fuller and Cynthia Gao and Vedanuj Goswami and Naman Goyal and Anthony Hartshorn and Saghar Hosseini and Rui Hou and Hakan Inan and Marcin Kardas and Viktor Kerkez and Madian Khabsa and Isabel Kloumann and Artem Korenev and Punit Singh Koura and Marie-Anne Lachaux and Thibaut Lavril and Jenya Lee and Diana Liskovich and Yinghai Lu and Yuning Mao and Xavier Martinet and Todor Mihaylov and Pushkar Mishra and Igor Molybog and Yixin Nie and Andrew Poulton and Jeremy Reizenstein and Rashi Rungta and Kalyan Saladi and Alan Schelten and Ruan Silva and Eric Michael Smith and Ranjan Subramanian and Xiaoqing Ellen Tan and Binh Tang and Ross Taylor and Adina Williams and Jian Xiang Kuan and Puxin Xu and Zheng Yan and Iliyan Zarov and Yuchen Zhang and Angela Fan and Melanie Kambadur and Sharan Narang and Aurelien Rodriguez and Robert Stojnic and Sergey Edunov and Thomas Scialom},
      year={2023},
      eprint={2307.09288},
      archivePrefix={arXiv},
      primaryClass={cs.CL},
      url={https://arxiv.org/abs/2307.09288}, 
}

@misc{openai2023gpt35turbo,
  author       = {{OpenAI}},
  title        = {GPT-3.5 Turbo},
  year         = {2023},
  howpublished = {OpenAI API Documentation},
  url          = {https://platform.openai.com/docs/models/gpt-3.5-turbo},
  note         = {Accessed: 2025-12-26}
}

@misc{openai2024gpt4o,
  author       = {{OpenAI}},
  title        = {GPT-4o},
  year         = {2024},
  howpublished = {OpenAI API Documentation},
  url          = {https://platform.openai.com/docs/models/gpt-4o},
  note         = {Accessed: 2025-12-26}
}

@misc{openai2024gpt4omini,
  author       = {{OpenAI}},
  title        = {GPT-4o mini},
  year         = {2024},
  howpublished = {OpenAI API documentation},
  url          = {https://platform.openai.com/docs/models/gpt-4o-mini},
  note         = {Accessed: 2025-12-26}
}

@misc{openai2024o1,
  author       = {{OpenAI}},
  title        = {{o1}},
  year         = {2024},
  howpublished = {OpenAI API Documentation},
  url          = {https://platform.openai.com/docs/models/o1},
  note         = {Accessed: 2025-12-26}
}

@article{deepseek2025r1,
   title={DeepSeek-R1 incentivizes reasoning in LLMs through reinforcement learning},
   volume={645},
   ISSN={1476-4687},
   url={http://dx.doi.org/10.1038/s41586-025-09422-z},
   DOI={10.1038/s41586-025-09422-z},
   number={8081},
   journal={Nature},
   publisher={Springer Science and Business Media LLC},
   author={Guo, Daya and Yang, Dejian and Zhang, Haowei and Song, Junxiao and Wang, Peiyi and Zhu, Qihao and Xu, Runxin and Zhang, Ruoyu and Ma, Shirong and Bi, Xiao and Zhang, Xiaokang and Yu, Xingkai and Wu, Yu and Wu, Z. F. and Gou, Zhibin and Shao, Zhihong and Li, Zhuoshu and Gao, Ziyi and Liu, Aixin and Xue, Bing and Wang, Bingxuan and Wu, Bochao and Feng, Bei and Lu, Chengda and Zhao, Chenggang and Deng, Chengqi and Ruan, Chong and Dai, Damai and Chen, Deli and Ji, Dongjie and Li, Erhang and Lin, Fangyun and Dai, Fucong and Luo, Fuli and Hao, Guangbo and Chen, Guanting and Li, Guowei and Zhang, H. and Xu, Hanwei and Ding, Honghui and Gao, Huazuo and Qu, Hui and Li, Hui and Guo, Jianzhong and Li, Jiashi and Chen, Jingchang and Yuan, Jingyang and Tu, Jinhao and Qiu, Junjie and Li, Junlong and Cai, J. L. and Ni, Jiaqi and Liang, Jian and Chen, Jin and Dong, Kai and Hu, Kai and You, Kaichao and Gao, Kaige and Guan, Kang and Huang, Kexin and Yu, Kuai and Wang, Lean and Zhang, Lecong and Zhao, Liang and Wang, Litong and Zhang, Liyue and Xu, Lei and Xia, Leyi and Zhang, Mingchuan and Zhang, Minghua and Tang, Minghui and Zhou, Mingxu and Li, Meng and Wang, Miaojun and Li, Mingming and Tian, Ning and Huang, Panpan and Zhang, Peng and Wang, Qiancheng and Chen, Qinyu and Du, Qiushi and Ge, Ruiqi and Zhang, Ruisong and Pan, Ruizhe and Wang, Runji and Chen, R. J. and Jin, R. L. and Chen, Ruyi and Lu, Shanghao and Zhou, Shangyan and Chen, Shanhuang and Ye, Shengfeng and Wang, Shiyu and Yu, Shuiping and Zhou, Shunfeng and Pan, Shuting and Li, S. S. and Zhou, Shuang and Wu, Shaoqing and Yun, Tao and Pei, Tian and Sun, Tianyu and Wang, T. and Zeng, Wangding and Liu, Wen and Liang, Wenfeng and Gao, Wenjun and Yu, Wenqin and Zhang, Wentao and Xiao, W. L. and An, Wei and Liu, Xiaodong and Wang, Xiaohan and Chen, Xiaokang and Nie, Xiaotao and Cheng, Xin and Liu, Xin and Xie, Xin and Liu, Xingchao and Yang, Xinyu and Li, Xinyuan and Su, Xuecheng and Lin, Xuheng and Li, X. Q. and Jin, Xiangyue and Shen, Xiaojin and Chen, Xiaosha and Sun, Xiaowen and Wang, Xiaoxiang and Song, Xinnan and Zhou, Xinyi and Wang, Xianzu and Shan, Xinxia and Li, Y. K. and Wang, Y. Q. and Wei, Y. X. and Zhang, Yang and Xu, Yanhong and Li, Yao and Zhao, Yao and Sun, Yaofeng and Wang, Yaohui and Yu, Yi and Zhang, Yichao and Shi, Yifan and Xiong, Yiliang and He, Ying and Piao, Yishi and Wang, Yisong and Tan, Yixuan and Ma, Yiyang and Liu, Yiyuan and Guo, Yongqiang and Ou, Yuan and Wang, Yuduan and Gong, Yue and Zou, Yuheng and He, Yujia and Xiong, Yunfan and Luo, Yuxiang and You, Yuxiang and Liu, Yuxuan and Zhou, Yuyang and Zhu, Y. X. and Huang, Yanping and Li, Yaohui and Zheng, Yi and Zhu, Yuchen and Ma, Yunxian and Tang, Ying and Zha, Yukun and Yan, Yuting and Ren, Z. Z. and Ren, Zehui and Sha, Zhangli and Fu, Zhe and Xu, Zhean and Xie, Zhenda and Zhang, Zhengyan and Hao, Zhewen and Ma, Zhicheng and Yan, Zhigang and Wu, Zhiyu and Gu, Zihui and Zhu, Zijia and Liu, Zijun and Li, Zilin and Xie, Ziwei and Song, Ziyang and Pan, Zizheng and Huang, Zhen and Xu, Zhipeng and Zhang, Zhongyu and Zhang, Zhen},
   year={2025},
   month=sep, pages={633–638} }

@inproceedings{huang-etal-2025-empirical,
    title = "An Empirical Study of {LLM}-as-a-Judge for {LLM} Evaluation: Fine-tuned Judge Model is not a General Substitute for {GPT}-4",
    author = "Huang, Hui  and
      Bu, Xingyuan  and
      Zhou, Hongli  and
      Qu, Yingqi  and
      Liu, Jing  and
      Yang, Muyun  and
      Xu, Bing  and
      Zhao, Tiejun",
    editor = "Che, Wanxiang  and
      Nabende, Joyce  and
      Shutova, Ekaterina  and
      Pilehvar, Mohammad Taher",
    booktitle = "Findings of the Association for Computational Linguistics: ACL 2025",
    month = jul,
    year = "2025",
    address = "Vienna, Austria",
    publisher = "Association for Computational Linguistics",
    url = "https://aclanthology.org/2025.findings-acl.306/",
    doi = "10.18653/v1/2025.findings-acl.306",
    pages = "5880--5895",
    ISBN = "979-8-89176-256-5"
}

@misc{liu2019roberta,
      title={RoBERTa: A Robustly Optimized BERT Pretraining Approach}, 
      author={Yinhan Liu and Myle Ott and Naman Goyal and Jingfei Du and Mandar Joshi and Danqi Chen and Omer Levy and Mike Lewis and Luke Zettlemoyer and Veselin Stoyanov},
      year={2019},
      eprint={1907.11692},
      archivePrefix={arXiv},
      primaryClass={cs.CL},
      url={https://arxiv.org/abs/1907.11692}, 
}

@inproceedings{
he2023debertav,
title={De{BERT}aV3: Improving De{BERT}a using {ELECTRA}-Style Pre-Training with Gradient-Disentangled Embedding Sharing},
author={Pengcheng He and Jianfeng Gao and Weizhu Chen},
booktitle={The Eleventh International Conference on Learning Representations },
year={2023},
url={https://openreview.net/forum?id=sE7-XhLxHA}
}

@inproceedings{lewis-etal-2020-bart,
    title = "{BART}: Denoising Sequence-to-Sequence Pre-training for Natural Language Generation, Translation, and Comprehension",
    author = "Lewis, Mike  and
      Liu, Yinhan  and
      Goyal, Naman  and
      Ghazvininejad, Marjan  and
      Mohamed, Abdelrahman  and
      Levy, Omer  and
      Stoyanov, Veselin  and
      Zettlemoyer, Luke",
    editor = "Jurafsky, Dan  and
      Chai, Joyce  and
      Schluter, Natalie  and
      Tetreault, Joel",
    booktitle = "Proceedings of the 58th Annual Meeting of the Association for Computational Linguistics",
    month = jul,
    year = "2020",
    address = "Online",
    publisher = "Association for Computational Linguistics",
    url = "https://aclanthology.org/2020.acl-main.703/",
    doi = "10.18653/v1/2020.acl-main.703",
    pages = "7871--7880"
}

@misc{liu2025mindstepbystep,
      title={Mind Your Step (by Step): Chain-of-Thought can Reduce Performance on Tasks where Thinking Makes Humans Worse}, 
      author={Ryan Liu and Jiayi Geng and Addison J. Wu and Ilia Sucholutsky and Tania Lombrozo and Thomas L. Griffiths},
      year={2025},
      eprint={2410.21333},
      archivePrefix={arXiv},
      primaryClass={cs.LG},
      url={https://arxiv.org/abs/2410.21333}, 
}

@inproceedings{wang-etal-2025-ranked,
    title = "Ranked Voting based Self-Consistency of Large Language Models",
    author = "Wang, Weiqin  and
      Wang, Yile  and
      Huang, Hui",
    editor = "Che, Wanxiang  and
      Nabende, Joyce  and
      Shutova, Ekaterina  and
      Pilehvar, Mohammad Taher",
    booktitle = "Findings of the Association for Computational Linguistics: ACL 2025",
    month = jul,
    year = "2025",
    address = "Vienna, Austria",
    publisher = "Association for Computational Linguistics",
    url = "https://aclanthology.org/2025.findings-acl.744/",
    doi = "10.18653/v1/2025.findings-acl.744",
    pages = "14410--14426",
    ISBN = "979-8-89176-256-5"
}

@inproceedings{10.5555/3495724.3495977,
 author = {Stiennon, Nisan and Ouyang, Long and Wu, Jeffrey and Ziegler, Daniel and Lowe, Ryan and Voss, Chelsea and Radford, Alec and Amodei, Dario and Christiano, Paul F},
 booktitle = {Advances in Neural Information Processing Systems},
 editor = {H. Larochelle and M. Ranzato and R. Hadsell and M.F. Balcan and H. Lin},
 pages = {3008--3021},
 publisher = {Curran Associates, Inc.},
 title = {Learning to summarize with human feedback},
 url = {https://proceedings.neurips.cc/paper_files/paper/2020/file/1f89885d556929e98d3ef9b86448f951-Paper.pdf},
 volume = {33},
 year = {2020}
}

@article{vashurin-etal-2025-benchmarking,
    title = "Benchmarking Uncertainty Quantification Methods for Large Language Models with {LM}-Polygraph",
    author = "Vashurin, Roman  and
      Fadeeva, Ekaterina  and
      Vazhentsev, Artem  and
      Rvanova, Lyudmila  and
      Vasilev, Daniil  and
      Tsvigun, Akim  and
      Petrakov, Sergey  and
      Xing, Rui  and
      Sadallah, Abdelrahman  and
      Grishchenkov, Kirill  and
      Panchenko, Alexander  and
      Baldwin, Timothy  and
      Nakov, Preslav  and
      Panov, Maxim  and
      Shelmanov, Artem",
    journal = "Transactions of the Association for Computational Linguistics",
    volume = "13",
    year = "2025",
    address = "Cambridge, MA",
    publisher = "MIT Press",
    url = "https://aclanthology.org/2025.tacl-1.11/",
    doi = "10.1162/tacl_a_00737",
    pages = "220--248"
}

@misc{kossen2024semantic,
      title={Semantic Entropy Probes: Robust and Cheap Hallucination Detection in LLMs}, 
      author={Jannik Kossen and Jiatong Han and Muhammed Razzak and Lisa Schut and Shreshth Malik and Yarin Gal},
      year={2024},
      eprint={2406.15927},
      archivePrefix={arXiv},
      primaryClass={cs.CL},
      url={https://arxiv.org/abs/2406.15927}, 
}

@inproceedings{
golovneva2023roscoe,
title={{ROSCOE}: A Suite of Metrics for Scoring Step-by-Step Reasoning},
author={Olga Golovneva and Moya Peng Chen and Spencer Poff and Martin Corredor and Luke Zettlemoyer and Maryam Fazel-Zarandi and Asli Celikyilmaz},
booktitle={The Eleventh International Conference on Learning Representations },
year={2023},
url={https://openreview.net/forum?id=xYlJRpzZtsY}
}

@inproceedings{chiang-lee-2024-reasoning,
    title = "Over-Reasoning and Redundant Calculation of Large Language Models",
    author = "Chiang, Cheng-Han  and
      Lee, Hung-yi",
    editor = "Graham, Yvette  and
      Purver, Matthew",
    booktitle = "Proceedings of the 18th Conference of the European Chapter of the Association for Computational Linguistics (Volume 2: Short Papers)",
    month = mar,
    year = "2024",
    address = "St. Julian{'}s, Malta",
    publisher = "Association for Computational Linguistics",
    url = "https://aclanthology.org/2024.eacl-short.15/",
    doi = "10.18653/v1/2024.eacl-short.15",
    pages = "161--169"
}

@inproceedings{jang-etal-2025-verbosity,
    title = "Verbosity-Aware Rationale Reduction: Sentence-Level Rationale Reduction for Efficient and Effective Reasoning",
    author = "Jang, Joonwon  and
      Kim, Jaehee  and
      Kweon, Wonbin  and
      Lee, Seonghyeon  and
      Yu, Hwanjo",
    editor = "Che, Wanxiang  and
      Nabende, Joyce  and
      Shutova, Ekaterina  and
      Pilehvar, Mohammad Taher",
    booktitle = "Findings of the Association for Computational Linguistics: ACL 2025",
    month = jul,
    year = "2025",
    address = "Vienna, Austria",
    publisher = "Association for Computational Linguistics",
    url = "https://aclanthology.org/2025.findings-acl.1068/",
    doi = "10.18653/v1/2025.findings-acl.1068",
    pages = "20769--20784",
    ISBN = "979-8-89176-256-5"
}

@inproceedings{
hendrycks2021measuringmath,
title={Measuring Mathematical Problem Solving With the {MATH} Dataset},
author={Dan Hendrycks and Collin Burns and Saurav Kadavath and Akul Arora and Steven Basart and Eric Tang and Dawn Song and Jacob Steinhardt},
booktitle={Thirty-fifth Conference on Neural Information Processing Systems Datasets and Benchmarks Track (Round 2)},
year={2021},
url={https://openreview.net/forum?id=7Bywt2mQsCe}
}

@inproceedings{
yao2023react,
title={ReAct: Synergizing Reasoning and Acting in Language Models},
author={Shunyu Yao and Jeffrey Zhao and Dian Yu and Nan Du and Izhak Shafran and Karthik R Narasimhan and Yuan Cao},
booktitle={The Eleventh International Conference on Learning Representations },
year={2023},
url={https://openreview.net/forum?id=WE_vluYUL-X}
}

@inproceedings{
shinn2023reflexion,
title={Reflexion: language agents with verbal reinforcement learning},
author={Noah Shinn and Federico Cassano and Ashwin Gopinath and Karthik R Narasimhan and Shunyu Yao},
booktitle={Thirty-seventh Conference on Neural Information Processing Systems},
year={2023},
url={https://openreview.net/forum?id=vAElhFcKW6}
}

@article{Hu_Dong_Sun_Huang_2026, title={Tapas Are Free! Training-Free Adaptation of Programmatic Agents via LLM-Guided Program Synthesis in Dynamic Environments}, volume={40}, url={https://ojs.aaai.org/index.php/AAAI/article/view/40189}, DOI={10.1609/aaai.v40i35.40189}, abstractNote={Autonomous agents in safety-critical applications must continuously adapt to dynamic conditions without compromising performance and reliability. This work introduces TAPA (Training-free Adaptation of Programmatic Agents), a novel framework that positions large language models (LLMs) as intelligent moderators of the symbolic action space. Unlike prior programmatic agents typically generate a monolithic policy program or rely on fixed symbolic action sets, TAPA synthesizes and adapts modular programs for individual high-level actions, referred to as logical primitives. By decoupling strategic intent from execution, TAPA enables meta-agents to operate over an abstract, interpretable action space while the LLM dynamically generates, composes, and refines symbolic programs tailored to each primitive. Extensive experiments across cybersecurity and swarm intelligence domains validate TAPA’s effectiveness. In autonomous DDoS defense scenarios, TAPA achieves 77.7% network uptime while maintaining near-perfect detection accuracy in unknown dynamic environments. In swarm intelligence formation control under environmental and adversarial disturbances, TAPA consistently preserves consensus at runtime where baseline methods fail. This work promotes a paradigm shift for autonomous system design in evolving environments, from policy adaptation to dynamic action adaptation.}, number={35}, journal={Proceedings of the AAAI Conference on Artificial Intelligence}, author={Hu, Jinwei and Dong, Yi and Sun, Youcheng and Huang, Xiaowei}, year={2026}, month={Mar.}, pages={29477-29485} }

@misc{hu2026lyingtruthsopenchannelmultiagent,
      title={Lying with Truths: Open-Channel Multi-Agent Collusion for Belief Manipulation via Generative Montage}, 
      author={Jinwei Hu and Xinmiao Huang and Youcheng Sun and Yi Dong and Xiaowei Huang},
      year={2026},
      eprint={2601.01685},
      archivePrefix={arXiv},
      primaryClass={cs.CL},
      url={https://arxiv.org/abs/2601.01685}, 
}

@misc{hu2025stopreducingresponsibilityllmpowered,
      title={Stop Reducing Responsibility in LLM-Powered Multi-Agent Systems to Local Alignment}, 
      author={Jinwei Hu and Yi Dong and Shuang Ao and Zhuoyun Li and Boxuan Wang and Lokesh Singh and Guangliang Cheng and Sarvapali D. Ramchurn and Xiaowei Huang},
      year={2025},
      eprint={2510.14008},
      archivePrefix={arXiv},
      primaryClass={cs.MA},
      url={https://arxiv.org/abs/2510.14008}, 
}

@inproceedings{bai-etal-2024-mt,
    title = "{MT}-Bench-101: A Fine-Grained Benchmark for Evaluating Large Language Models in Multi-Turn Dialogues",
    author = "Bai, Ge  and
      Liu, Jie  and
      Bu, Xingyuan  and
      He, Yancheng  and
      Liu, Jiaheng  and
      Zhou, Zhanhui  and
      Lin, Zhuoran  and
      Su, Wenbo  and
      Ge, Tiezheng  and
      Zheng, Bo  and
      Ouyang, Wanli",
    editor = "Ku, Lun-Wei  and
      Martins, Andre  and
      Srikumar, Vivek",
    booktitle = "Proceedings of the 62nd Annual Meeting of the Association for Computational Linguistics (Volume 1: Long Papers)",
    month = aug,
    year = "2024",
    address = "Bangkok, Thailand",
    publisher = "Association for Computational Linguistics",
    url = "https://aclanthology.org/2024.acl-long.401/",
    doi = "10.18653/v1/2024.acl-long.401",
    pages = "7421--7454"
}

@inproceedings{liu-etal-2023-g,
    title = "{G}-Eval: {NLG} Evaluation using Gpt-4 with Better Human Alignment",
    author = "Liu, Yang  and
      Iter, Dan  and
      Xu, Yichong  and
      Wang, Shuohang  and
      Xu, Ruochen  and
      Zhu, Chenguang",
    editor = "Bouamor, Houda  and
      Pino, Juan  and
      Bali, Kalika",
    booktitle = "Proceedings of the 2023 Conference on Empirical Methods in Natural Language Processing",
    month = dec,
    year = "2023",
    address = "Singapore",
    publisher = "Association for Computational Linguistics",
    url = "https://aclanthology.org/2023.emnlp-main.153/",
    doi = "10.18653/v1/2023.emnlp-main.153",
    pages = "2511--2522"
}

\appendix
\section{CoT Prompting Template}
\label{app:prompts}
For all models, we use a two-part prompt consisting
of a system message and a user message. Below
we list the exact templates and hyperparameter settings used in our experiments.

\subsection{Base CoT Prompt with Reference}

\begin{promptbox}[System message]
You are a careful reasoning assistant.
Output ONLY one <json>{...}</json> block, no extra text.
\end{promptbox}

\begin{promptbox}[User message]
Here is a worked example of multiple-choice reasoning with a concise chain.

Example Question: {RQ}
Example Context: {RCTX}
Example Options: (A) {ROPT_A} (B) {ROPT_B} (C) {ROPT_C} (D) {ROPT_D}
Example Answer: {RANS}
Example Chain (strict JSON):
<json>{"because_steps": ["STEP_1", ..., "STEP_L"],
       "therefore": "FINAL_SENTENCE",
       "answer": "{RANS}"}</json>

Now solve the NEW problem in the SAME STYLE and EXACTLY the SAME NUMBER OF STEPS.
- Produce EXACTLY {L_steps} steps in "because_steps".
- Keep each step to ONE short sentence; keep entities consistent; avoid hedging.
- Output ONLY one
  <json>{"because_steps":["..."],"therefore":"...","answer":"A"}</json> block. No extra text.
- The field "answer" MUST be one of A, B, C, or D and MUST be a single letter (e.g., "A").

Question: {TQ}
Context: {TCTX}
Options: (A) {TOPT_A} (B) {TOPT_B} (C) {TOPT_C} (D) {TOPT_D}
\end{promptbox}

Here, \verb|{RQ,RCTX,ROPT_*,RANS}| come from the reference chain,
\verb|{TQ,TCTX,TOPT_*}| from the target QA instance, and
\verb|{L_steps}| is the hop number $L\in\{1,2,3,4\}$.

\subsection{Hyperparameter Settings}
\label{subsec:hparams}

\paragraph{OpenAI chat models.}
Unless otherwise specified, GPT-4o and GPT-4o-mini are called with temperature $=0.4$, top-$p=0.9$, \texttt{n=1}, and a maximum generation length of 640 tokens. For multi-sample settings (SC-CoT, SC-Align with multiple candidate chains), we increase the temperature to $0.7$ while keeping top-$p=0.9$ and the same length.

\paragraph{Open-source models.}
Open-source LLMs are queried via the HuggingFace \texttt{text-generation} pipeline with \texttt{do\_sample=true}, using the same temperature and top-$p$ values as their OpenAI counterparts in each setting, and the same maximum length (640 tokens).

\paragraph{Repetitions and reporting.}
Each configuration (model, hop, selection strategy) is repeated over 5 random seeds. We report the mean and standard deviation of the Accuracy, the Alignment Score and LLM-based evaluation scores across these runs.

\section{Threshold Selection for \texorpdfstring{$\tau_{\mathrm{TS}}$}{tau_TS} and \texorpdfstring{$\tau_{\mathrm{RR}}$}{tau_RR}}
\label{app:threshold}

\begin{table}[t]
\centering
\setlength{\tabcolsep}{4.0pt}
\renewcommand{\arraystretch}{1.08}

\label{tab:threshold_summary}
\begin{tabular}{@{}lcc@{}}
\toprule
Metric & Percentile $q$ & Threshold \\
\midrule
TS & 5  & $\tau_{\mathrm{TS}}=0.425$ \\
TS & 10 & $\tau_{\mathrm{TS}}=0.563$ \\
TS & 15 & $\tau_{\mathrm{TS}}=0.712$ \\
\midrule
RR & 70 & $\tau_{\mathrm{RR}}=0.794$ \\
RR & 75 & $\tau_{\mathrm{RR}}=0.816$ \\
RR & 80 & $\tau_{\mathrm{RR}}=0.858$ \\
\bottomrule
\end{tabular}
\caption{Threshold summary for TS and RR using percentile-based calibration.}
\label{tab:appendix_threshold}
\end{table}

We choose the TS/RR thresholds via percentile calibration to avoid hand-tuning absolute cosine cutoffs, which can vary with the embedding model and prompting format.
As detailed in Table~\ref{tab:appendix_threshold}, for thematic shift (TS), we target \emph{unusually low} semantic continuity between adjacent steps; thus $\tau_{\mathrm{TS}}$ is set from the \emph{lower tail} (a small percentile) of adjacent-step similarities $\{\cos(e_{B_i}, e_{B_{i+1}})\}$.
For redundant reasoning (RR), we target \emph{unusually high} semantic overlap with earlier steps; thus $\tau_{\mathrm{RR}}$ is set from the \emph{upper tail} (a high percentile) of $\{\max_{j<i}\cos(e_{B_i}, e_{B_j})\}$.
Unless otherwise stated, we instantiate these percentiles as $q_{\mathrm{TS}}{=}10$ and $q_{\mathrm{RR}}{=}75$, yielding $\tau_{\mathrm{TS}}=Q_{10}(\{\cos(e_{B_i}, e_{B_{i+1}})\})$ and $\tau_{\mathrm{RR}}=Q_{75}(\{\max_{j<i}\cos(e_{B_i}, e_{B_j})\})$, where $Q_q(\cdot)$ denotes the $q$-th percentile.
We compute these percentiles on a development split constructed from 256 sets of 4-hop reasoning chains generated by LLaMA-2-13B-Chat, which provides a reference distribution for calibrating semantic continuity and redundancy.

\section{LLM-based Evaluation}
\label{app:llm-eval-template}

\subsection{Prompt Template}

We use an LLM-as-judge protocol to compare the baseline chain with the SC-Align
chain for the same multiple-choice question.
For each matched pair $(\text{Hop}, \text{Idx})$, we construct two textual
chains and randomly decide which one is shown as ``Chain 1'' and which as
``Chain 2''.
The judge model evaluates both chains along two
dimensions:

\begin{itemize}
    \item \textbf{Logical completeness}: whether the chain covers the key
    reasoning steps needed to justify the answer and maintains coherent logic.
    \item \textbf{Readability}: whether the chain is easy to understand and free of confusing repetition.
\end{itemize}

The judge assigns integer scores in $[1,10]$ for each dimension and chain,
decides which chain is better (or a tie) on each dimension, and provides a brief
textual explanation. The order of chains is stored in the metadata so that
scores can be mapped back to baseline vs.\ SC-Align.
\begin{promptbox}[System message (judge)]
You are a strict but fair evaluator of
reasoning quality.
    
\end{promptbox}
\begin{promptbox}[User message (judge)]
You are an expert evaluator of reasoning chains.
You will be given:
- A multiple-choice question and its options.
- Two reasoning chains that both try to answer this question:
  Chain 1 is produced by **{CHAIN1_NAME}**.
  Chain 2 is produced by **{CHAIN2_NAME}**.
You must judge each chain along two dimensions:
1. Logical completeness: Does the chain cover the key reasoning steps needed
   to justify the answer? Is the causal logic coherent and sufficiently detailed?
2. Readability: Is the chain easy to understand, well-structured, and free of
   confusing repetition?
For EACH dimension, assign a score from 1 (very poor) to 10 (excellent) to BOTH
chains. Then decide which chain is better on that dimension (or "tie" if they
are comparable).
Return your judgment as a JSON object with the following fields ONLY:
{
  "chain1_logic": <integer 1-10>,
  "chain2_logic": <integer 1-10>,
  "chain1_readability": <integer 1-10>,
  "chain2_readability": <integer 1-10>,
  "better_logic": "chain1" | "chain2" | "tie",
  "better_readability": "chain1" | "chain2" | "tie",
  "explanation": "<brief explanation in 1-3 sentences>"
}
Make sure the output is valid JSON (no comments, no trailing commas).
Question:
{QUESTION}
Options:
{OPTION_KEY}. {OPTION_TEXT}
...
--- Chain 1 ---
{CHAIN1_TEXT}
--- Chain 2 ---
{CHAIN2_TEXT}
\end{promptbox}

\section{Construction of Reference Chains}
\label{app:ref-chains}

To obtain high-quality reference chains for each reasoning hop
$L \in \{1,2,3,4\}$, we apply an AI-assisted generation pipeline with LLM-based evaluation. Each
reference chain contains exactly $L$ intermediate reasoning steps followed
by a final conclusion and is stored in JSON format.

\paragraph{Candidate generation.}
For each question and hop $L$, we generate candidate CoT chains using a
pool of strong LLMs. For each sample, we randomly choose one model from
GPT-5, GPT-4.1, and Claude 3.5 Sonnet to generate the chain. These
generators are different from the models evaluated in our experiments. This
stage produces more than 15,000 candidate chains in total.

\paragraph{Correctness filtering.}
We discard all candidate chains whose predicted answers do not match the
gold label. This step ensures that the reference pool contains only chains
that lead to the correct final answer, before any LLM-based filtering is applied.

\paragraph{LLM-as-a-judge pre-filtering.}
To reduce the amount of manual check required, we use an LLM judge
following Appendix~\ref{app:llm-eval-template}. The judge scores each
remaining chain on \emph{readability} and \emph{logical coherence}, both on
a 1--10 scale. We retain only chains with an average score greater than
8.50, which reduces the pool from more than 15,000 candidates to roughly 2,000
high-quality chains.

\paragraph{Author-side sanity check.}
Finally, we conduct an internal author-side sanity check for quality
assurance, rather than collecting human preference labels. In small batches,
we inspect whether the selected chains are readable, logically coherent,
non-redundant, and free from generic meta-commentary such as ``I will now
think step by step.'' After this process, we obtain 256 reference chains for
each reasoning hop from 1-hop to 4-hop, resulting in 1,024 reference chains
in total.

\definecolor{cotblock}{RGB}{245,245,245}    
\definecolor{cssblock}{RGB}{238,245,252}    
\definecolor{alignblock}{RGB}{250,244,234}  

\begin{table}[t]
\centering
\small
\setlength{\tabcolsep}{1pt}
\renewcommand{\arraystretch}{1.15}

\newcommand{\mth}[1]{\makebox[3.2em][l]{\scriptsize #1}}

\begin{tabular}{@{}l c c c@{}}
\toprule

& \multicolumn{1}{c}{\textbf{Open-Source Models}}
& \multicolumn{2}{c}{\textbf{Closed-Source Models}} \\
\cmidrule(lr){2-2} \cmidrule(lr){3-4}

\textbf{Method}
& \scriptsize\textbf{Qwen3-4B-Instruct}
& \scriptsize\textbf{GPT-5-mini}
& \scriptsize\textbf{GPT-5} \\
\midrule

\multicolumn{4}{c}{\textit{Accuracy (mean $\pm$ std, \%)}} \\
\midrule

\rowcolor{cotblock}
\mth{CoT}
& $89.30\;\text{\scriptsize($\pm$2.46)}$
& $95.67\;\text{\scriptsize($\pm$1.21)}$
& $94.47\;\text{\scriptsize($\pm$0.89)}$ \\

\rowcolor{cotblock}
\mth{SC-CoT}
& $\mathbf{\textcolor{blue}{90.86}\;\text{\scriptsize($\pm$2.85)}}$
& $95.62\;\text{\scriptsize($\pm$0.37)}$
& $\mathbf{\textcolor{red}{96.81}\;\text{\scriptsize($\pm$2.18)}}$ \\

\rowcolor{cssblock}
\mth{ACSS-TS}
& $89.35\;\text{\scriptsize($\pm$2.82)}$
& $95.55\;\text{\scriptsize($\pm$1.72)}$
& $95.38\;\text{\scriptsize($\pm$1.58)}$ \\

\rowcolor{cssblock}
\mth{ACSS-RR}
& $89.22\;\text{\scriptsize($\pm$4.94)}$
& $95.43\;\text{\scriptsize($\pm$1.20)}$
& $95.70\;\text{\scriptsize($\pm$2.38)}$ \\

\rowcolor{cssblock}
\mth{ACSS-Ali.}
& $\mathbf{\textcolor{red}{91.13}\;\text{\scriptsize($\pm$2.72)}}$
& $\mathbf{\textcolor{red}{96.41}\;\text{\scriptsize($\pm$0.82)}}$
& $95.84\;\text{\scriptsize($\pm$2.67)}$ \\

\rowcolor{alignblock}
\mth{SC-Align}
& $90.38\;\text{\scriptsize($\pm$2.56)}$
& $\mathbf{\textcolor{blue}{95.78}\;\text{\scriptsize($\pm$2.23)}}$
& $\mathbf{\textcolor{blue}{96.72}\;\text{\scriptsize($\pm$1.80)}}$ \\

\midrule
\multicolumn{4}{c}{\textit{Alignment Score (mean $\pm$ std)}} \\
\midrule

\rowcolor{cotblock}
\mth{CoT}
& $78.37\;\text{\scriptsize($\pm$13.07)}$
& $78.57\;\text{\scriptsize($\pm$12.95)}$
& $78.57\;\text{\scriptsize($\pm$13.31)}$ \\

\rowcolor{cotblock}
\mth{SC-CoT}
& $78.48\;\text{\scriptsize($\pm$13.41)}$
& $79.00\;\text{\scriptsize($\pm$13.13)}$
& $78.84\;\text{\scriptsize($\pm$12.90)}$ \\

\rowcolor{cssblock}
\mth{ACSS-TS}
& $78.30\;\text{\scriptsize($\pm$13.10)}$
& $79.33\;\text{\scriptsize($\pm$12.26)}$
& $79.45\;\text{\scriptsize($\pm$13.48)}$ \\

\rowcolor{cssblock}
\mth{ACSS-RR}
& $78.39\;\text{\scriptsize($\pm$13.37)}$
& $78.75\;\text{\scriptsize($\pm$13.37)}$
& $79.07\;\text{\scriptsize($\pm$12.87)}$ \\

\rowcolor{cssblock}
\mth{ACSS-Ali.}
& $\mathbf{\textcolor{blue}{84.23}\;\text{\scriptsize($\pm$11.25)}}$
& $\mathbf{\textcolor{red}{88.02}\;\text{\scriptsize($\pm$8.12)}}$
& $\mathbf{\textcolor{blue}{87.57}\;\text{\scriptsize($\pm$8.22)}}$ \\

\rowcolor{alignblock}
\mth{SC-Align}
& $\mathbf{\textcolor{red}{84.94}\;\text{\scriptsize($\pm$10.25)}}$
& $\mathbf{\textcolor{blue}{87.74}\;\text{\scriptsize($\pm$8.09)}}$
& $\mathbf{\textcolor{red}{88.00}\;\text{\scriptsize($\pm$8.10)}}$ \\

\bottomrule
\end{tabular}
\caption{Additional experiments under the same setup as the main paper. \textbf{\textcolor{red}{Red}} numbers indicate the best result, and \textbf{\textcolor{blue}{blue}} numbers indicate the second best.}
\label{tab:additional_models}
\end{table}

\section{Additional Experiments}
\label{app:additional_models}

To further test the generality of our findings, we evaluate three additional models, namely Qwen3-4B-Instruct, GPT-5-mini, and GPT-5, under the same setup as in the main paper. As shown in Table~\ref{tab:additional_models}, the overall trend remains consistent: improvements in the Alignment Score are closely associated with gains in task accuracy.

\section{Grid Ablation on LLM Judge Models}
\label{subsec:judge-ablation}

As shown in Table~\ref{tab:llm_judge_vertical_all_backbones}, we conduct a
\emph{grid ablation} over both the \textbf{LLM judge} and the \textbf{NLI backbone}. Specifically, we repeat the SC-CoT vs.\ SC-Align
comparison under three LLM judges (GPT-4o, GPT-4.1, and GPT-3.5-turbo) and three
NLI backbones (RoBERTa-large-MNLI, DeBERTa-v3-large-MNLI, and BART-large-MNLI),
forming a $3\times 3$ evaluation grid. Across all nine configurations, SC-Align
consistently achieves higher mean ratings in both \textit{coherence} and
\textit{readability}, and is preferred in pairwise comparisons by a clear margin.
The persistence of this advantage across diverse judges and backbone choices
suggests that the observed gains reflect a robust improvement in reasoning-chain
quality, rather than an artifact of any particular judge or NLI backbone.

\begin{table}[htbp]
\centering
\small
\setlength{\tabcolsep}{5pt}
\renewcommand{\arraystretch}{0.9}
\begin{tabular}{@{}p{1.3cm} l c c@{}}
\toprule
\textbf{Aspect} & \textbf{Method} & \textbf{Score (mean$\pm$std)} & \textbf{Pairwise} \\
\midrule

\rowcolor{gray!10}
\multicolumn{4}{c}{\textbf{NLI Backbone: RoBERTa-large-MNLI}} \\
\cmidrule(lr){1-4}

\rowcolor{gray!15}
\multicolumn{4}{c}{\textit{Judge Model: GPT-4o}} \\
\cmidrule(lr){1-4}
\multirow{2}{*}{Coherence}
  & SC-CoT   & 8.85{\scriptsize($\pm 0.71$)} & 16.1\% win \\
  & SC-Align & \textbf{8.91}{\scriptsize($\pm 0.76$)} & \textbf{23.9\% win} \\
\midrule
\multirow{2}{*}{Readability}
  & SC-CoT   & 8.65{\scriptsize($\pm 0.56$)} & 14.5\% win \\
  & SC-Align & \textbf{8.87}{\scriptsize($\pm 0.49$)} & \textbf{37.3\% win} \\
\midrule

\rowcolor{gray!15}
\multicolumn{4}{c}{\textit{Judge Model: GPT-4.1}} \\
\cmidrule(lr){1-4}
\multirow{2}{*}{Coherence}
  & SC-CoT   & 8.55{\scriptsize($\pm 0.95$)} & 35.7\% win \\
  & SC-Align & \textbf{8.70}{\scriptsize($\pm 0.99$)} & \textbf{53.3\% win} \\
\midrule
\multirow{2}{*}{Readability}
  & SC-CoT   & 8.80{\scriptsize($\pm 0.75$)} & 27.5\% win \\
  & SC-Align & \textbf{9.07}{\scriptsize($\pm 0.75$)} & \textbf{51.8\% win} \\
\midrule

\rowcolor{gray!15}
\multicolumn{4}{c}{\textit{Judge Model: GPT-3.5-turbo}} \\
\cmidrule(lr){1-4}
\multirow{2}{*}{Coherence}
  & SC-CoT   & 8.37{\scriptsize($\pm 0.58$)} & 23.1\% win \\
  & SC-Align & \textbf{8.60}{\scriptsize($\pm 0.61$)} & \textbf{44.7\% win} \\
\midrule
\multirow{2}{*}{Readability}
  & SC-CoT   & 8.06{\scriptsize($\pm 0.73$)} & 28.6\% win \\
  & SC-Align & \textbf{8.32}{\scriptsize($\pm 0.54$)} & \textbf{52.5\% win} \\
\midrule

\rowcolor{gray!10}
\multicolumn{4}{c}{\textbf{NLI Backbone: DeBERTa-v3-large-MNLI}} \\
\cmidrule(lr){1-4}

\rowcolor{gray!15}
\multicolumn{4}{c}{\textit{Judge Model: GPT-4o}} \\
\cmidrule(lr){1-4}
\multirow{2}{*}{Coherence}
  & SC-CoT   & 8.85{\scriptsize($\pm 0.75$)} & 14.1\% win \\
  & SC-Align & \textbf{8.95}{\scriptsize($\pm 0.69$)} & \textbf{23.5\% win} \\
\midrule
\multirow{2}{*}{Readability}
  & SC-CoT   & 8.69{\scriptsize($\pm 0.58$)} & 12.9\% win \\
  & SC-Align & \textbf{8.93}{\scriptsize($\pm 0.48$)} & \textbf{34.5\% win} \\
\midrule

\rowcolor{gray!15}
\multicolumn{4}{c}{\textit{Judge Model: GPT-4.1}} \\
\cmidrule(lr){1-4}
\multirow{2}{*}{Coherence}
  & SC-CoT   & 8.57{\scriptsize($\pm 0.90$)} & 36.5\% win \\
  & SC-Align & \textbf{8.76}{\scriptsize($\pm 0.89$)} & \textbf{52.2\% win} \\
\midrule
\multirow{2}{*}{Readability}
  & SC-CoT   & 8.86{\scriptsize($\pm 0.76$)} & 31.0\% win \\
  & SC-Align & \textbf{9.07}{\scriptsize($\pm 0.66$)} & \textbf{48.2\% win} \\
\midrule

\rowcolor{gray!15}
\multicolumn{4}{c}{\textit{Judge Model: GPT-3.5-turbo}} \\
\cmidrule(lr){1-4}
\multirow{2}{*}{Coherence}
  & SC-CoT   & 8.39{\scriptsize($\pm 0.58$)} & 26.3\% win \\
  & SC-Align & \textbf{8.54}{\scriptsize($\pm 0.67$)} & \textbf{40.0\% win} \\
\midrule
\multirow{2}{*}{Readability}
  & SC-CoT   & 8.11{\scriptsize($\pm 0.69$)} & 28.2\% win \\
  & SC-Align & \textbf{8.36}{\scriptsize($\pm 0.57$)} & \textbf{52.9\% win} \\
\midrule

\rowcolor{gray!10}
\multicolumn{4}{c}{\textbf{NLI Backbone: BART-large-MNLI}} \\
\cmidrule(lr){1-4}

\rowcolor{gray!15}
\multicolumn{4}{c}{\textit{Judge Model: GPT-4o}} \\
\cmidrule(lr){1-4}
\multirow{2}{*}{Coherence}
  & SC-CoT   & 8.83{\scriptsize($\pm 0.66$)} & 14.9\% win \\
  & SC-Align & \textbf{8.92}{\scriptsize($\pm 0.68$)} & \textbf{24.3\% win} \\
\midrule
\multirow{2}{*}{Readability}
  & SC-CoT   & 8.70{\scriptsize($\pm 0.53$)} & 14.9\% win \\
  & SC-Align & \textbf{8.91}{\scriptsize($\pm 0.49$)} & \textbf{36.1\% win} \\
\midrule

\rowcolor{gray!15}
\multicolumn{4}{c}{\textit{Judge Model: GPT-4.1}} \\
\cmidrule(lr){1-4}
\multirow{2}{*}{Coherence}
  & SC-CoT   & 8.55{\scriptsize($\pm 0.97$)} & 39.2\% win \\
  & SC-Align & \textbf{8.69}{\scriptsize($\pm 0.90$)} & \textbf{49.0\% win} \\
\midrule
\multirow{2}{*}{Readability}
  & SC-CoT   & 8.78{\scriptsize($\pm 0.71$)} & 31.0\% win \\
  & SC-Align & \textbf{9.00}{\scriptsize($\pm 0.74$)} & \textbf{51.4\% win} \\
\midrule

\rowcolor{gray!15}
\multicolumn{4}{c}{\textit{Judge Model: GPT-3.5-turbo}} \\
\cmidrule(lr){1-4}
\multirow{2}{*}{Coherence}
  & SC-CoT   & 8.35{\scriptsize($\pm 0.58$)} & 28.6\% win \\
  & SC-Align & \textbf{8.53}{\scriptsize($\pm 0.66$)} & \textbf{45.5\% win} \\
\midrule
\multirow{2}{*}{Readability}
  & SC-CoT   & 8.10{\scriptsize($\pm 0.70$)} & 30.2\% win \\
  & SC-Align & \textbf{8.36}{\scriptsize($\pm 0.56$)} & \textbf{54.5\% win} \\
\bottomrule
\end{tabular}
\caption{LLM-based evaluation of reasoning quality (SC-CoT vs.\ SC-Align) under three different LLM judges, across three NLI backbones used for Alignment Score calculation (RoBERTa-large-MNLI, DeBERTa-v3-large-MNLI, and BART-large-MNLI). Scores are on a 1--10 scale. \textbf{Bold} numbers indicate the higher values.}
\label{tab:llm_judge_vertical_all_backbones}
\end{table}

\section{Qualitative Case Study of Reasoning Alignment Comparison}
\label{app:qualitative-cases}

To complement the quantitative results, we conduct a qualitative inspection of the LLM-evaluation results and select representative SC-CoT vs.\ SC-Align pairs where the baseline chain exhibits a lower reasoning alignment and SC-Align exhibits a higher reasoning alignment according to our metric. In all cases, low-alignment baseline chains tend to (i) leave key relations implicit (missing “bridge” statements), (ii) repeat semantically similar points without making step-wise progress, or (iii) contain small local incoherence that reduces readability even when the conclusion is plausible. In contrast, SC-Align is often preferred because it states the intended causal/semantic links more explicitly and presents the steps in a more coherent and readable form. In the following, we summarize the results of the LLM-based evaluation and provide a qualitative analysis. We also include a failure case to illustrate a common error mode.

\paragraph{Case 1: Making implicit entity roles and properties explicit}
Here, the judge prefers SC-Align because it more explicitly distinguishes key entities and clarifies their roles and properties, resulting in a rationale that is both easier to follow and more logically coherent.

\begin{pro2}[SC-Align more explicitly distinguishes the two groups and improves logic and readability]
\textbf{Question:}
Some researchers believe that beta-carotene, which is found in carrots, may help humans live longer lives. A few scientists designed an experiment in which one-hundred people would be asked to take a double dose of beta-carotene daily for two years and another hundred people were given a placebo, a sugar pill, for the same amount of time. After the two years, the scientists planned to collect the data and draw a conclusion. This is not a good example of a scientific study and was not done because there is no control group. Is that correct?
\par\smallskip

\textbf{Baseline (SC-CoT; lower Alignment Score):}

Step 1: A control group is essential for comparing results against the experimental group.

Step 2: The placebo group serves as the control group in this study.

Step 3: Without a control group, it is difficult to determine the effect of beta-carotene.

Step 4: This study does have a control group, so that is not the issue.

Conclusion: The statement that there is no control group is incorrect.

\par\smallskip

\textbf{SC-Align (higher Alignment Score):}

Step 1: The study includes one group taking beta-carotene and another group taking a placebo.

Step 2: The group taking beta-carotene is the experimental group.

Step 3: The placebo group serves as the control group.

Step 4: Because the study includes a placebo control group, the absence of a control group is not the flaw in the design.

Conclusion: The statement that there is no control group is incorrect.

\par\smallskip

\textbf{LLM Judge Scores:}

Logic: Baseline vs SC-Align (better: SC-Align)

Readability: Baseline vs SC-Align (better: SC-Align)

\par\smallskip

\textbf{LLM Judge Explanation:}
SC-Align is preferred because it more explicitly distinguishes the two groups and states their roles more directly, making the reasoning easier to follow and logically clearer.
\end{pro2}

\paragraph{Case 2: Reducing redundancy while preserving the same reasoning trajectory.}
In this pair, both chains are logically correct, but SC-Align is preferred because it avoids minor repetition and uses more concise phrasing, improving readability without changing the underlying logic.

\begin{pro2}[Same chemistry but less repetition: SC-Align keeps the steps but removes redundancy]
\textbf{Question:}
Robert put two drops of an indicator into vinegar, and the color changed. Then he added ammonia solution until the color disappeared. What process occurred?

\par\medskip
\textbf{Baseline (SC-CoT; lower Alignment Score):}

Step 1: The indicator changes color in response to acidity or basicity.

Step 2: Vinegar is acidic, turning the indicator red.

Step 3: Adding ammonia, which is basic, neutralizes the acid in vinegar.

Step 4: The disappearance of red color indicates a neutralization reaction occurred.

Conclusion: The process that occurred is neutralization.

\par\medskip
\textbf{SC-Align (higher Alignment Score):}

Step 1: The indicator changes color in response to pH changes.

Step 2: Vinegar is acidic and turns the indicator red.

Step 3: Ammonia is a base that can neutralize acids.

Step 4: When the color disappears, it indicates a neutralization reaction occurred.

Conclusion: The process that occurred is neutralization.

\par\medskip
\textbf{LLM Judge Scores:}

Logic: Baseline vs SC-Align (better: tie)

Readability: Baseline vs SC-Align (better: SC-Align)

\par\medskip
\textbf{LLM Judge Explanation:}
Both chains logically explain the neutralization process. SC-Align is more readable due to its concise phrasing and lack of redundancy.
\end{pro2}

\paragraph{Case 3: Readability gains under the same logical content.}
This example illustrates a frequent pattern in our logs: the judge considers the logic comparable, but SC-Align is preferred because it is phrased more cleanly and reads more naturally.

\begin{pro2}[Equivalent causal story but clearer phrasing: readability improves with logic unchanged (logic tie)]
\textbf{Question:}
Which statement best explains why a tree branch floats on water?

\par\medskip
\textbf{Baseline (SC-CoT; lower Alignment Score):}

Step 1: Wood has air pockets that make it less dense than water.

Step 2: Because it is less dense, it will float on water.

Step 3: If an object is less dense than water, it will float.

Step 4: Therefore, wood is buoyant.

Conclusion: This means wood is buoyant.

\par\medskip
\textbf{SC-Align (higher Alignment Score):}

Step 1: Wood contains air-filled spaces that lower its overall density.

Step 2: Lower density than water allows an object to float.

Step 3: A floating object demonstrates buoyancy.

Step 4: Thus, the branch floats because wood is buoyant.

Conclusion: This means wood is buoyant.

\par\medskip
\textbf{LLM Judge Scores:}

Logic: Baseline vs SC-Align (better: tie)

Readability: Baseline vs SC-Align (better: SC-Align)

\par\medskip
\textbf{LLM Judge Explanation:}
Both chains provide a complete and coherent reasoning process. However, SC-Align is slightly more readable due to its clearer phrasing and structure.
\end{pro2}

\paragraph{Case 4 (Failure case): SC-Align introduces an extra assumption and is penalized.}
Finally, we include a counterexample where the baseline is preferred. This highlights a typical failure mode: SC-Align may add an unnecessary or less-supported assumption, which can reduce logical precision and hurt readability.

\begin{pro2}[Failure mode: added assumption reduces alignment (baseline preferred by the judge)]
\textbf{Question:}
Some animals live in a grassland. A fire burns through the grassland. Which type of animal will best compete for resources after the fire?

\par\medskip
\textbf{Baseline (SC-CoT; lower Alignment Score):}

Step 1: The fire destroys above-ground vegetation.

Step 2: Animals that rely on plants for shelter will struggle after the fire.

Step 3: Animals that live underground can avoid the immediate effects of the fire.

Step 4: Those animals can access resources that are less affected by the fire.

Conclusion: An animal that lives under the ground will best compete for resources after the fire.

\par\medskip
\textbf{SC-Align (higher Alignment Score):}

Step 1: The fire destroys above-ground vegetation.

Step 2: Animals that rely on plants for shelter will struggle after the fire.

Step 3: Animals that live underground can avoid the immediate effects of the fire.

Step 4: Animals that build nests on the ground may lose their nests and face resource competition.

Conclusion: An animal that lives under the ground will best compete for resources after the fire.

\par\medskip
\textbf{LLM Judge Scores:}

Logic: Baseline vs SC-Align (better: baseline)

Readability: Baseline vs SC-Align (better: baseline)

\par\medskip
\textbf{LLM Judge Explanation:}
Baseline is more logically complete because it directly connects underground living to accessing resources that remain available after the fire. SC-Align introduces extra details about nests that are not necessary for the conclusion, which slightly reduces clarity.
\end{pro2}

\paragraph{Summary.}
Overall, these case studies align with the intended behavior of Alignment Score: lower-alignment chains are often less coherent due to missing links, redundancy, or added assumptions, while higher-alignment SC-Align chains tend to be more coherent and more readable. The failure case further motivates using alignment-based diagnostics to detect when a rewrite adds unnecessary content or weakens the original reasoning structure.
\end{document}